\documentclass[a4paper,11pt,final]{article}

\usepackage{graphicx,amsmath,amssymb} 
\usepackage{epsfig}

\begin{document}

\title{\ \\ \LARGE\bf  Codynamic Fitness Landscapes of  Coevolutionary Minimal Substrates}

\author{Hendrik Richter\\
HTWK Leipzig University of Applied Sciences\\ Faculty of
Electrical Engineering and Information Technology\\
        Postfach 301166, D--04251 Leipzig, Germany. \\ Email: 
richter@eit.htwk-leipzig.de.  }

\maketitle

\begin{abstract}
Coevolutionary minimal substrates are simple and abstract models that allow  studying the  relationships and codynamics between objective and subjective fitness.  Using these models an approach  is presented for defining and analyzing fitness landscapes of coevolutionary problems. We devise similarity measures of codynamic fitness landscapes and experimentally study minimal substrates of  test--based and compositional problems for both cooperative and competitive interaction. 
\end{abstract}

\section{Introduction}

Coevolutionary scenarios are interesting for at least two reasons. A first is that in  natural evolution of biological entities the evolutionary development of one species is almost always accompanied by evolutionary adaption of and changes in other species.  Traits, features and abilities in one species do not exist for themselves, but can only be understood by the coupling with and response to other species' evolution.   Hence, studying natural evolution most likely means analyzing coevolutionary processes. A second reason is that in artificial evolution used to solve optimization problems by evolutionary search algorithms, designs employing ideas from coevolution appear to be as intriguing as promising. The advantages of coevolutionary designs are particularly seen for solving competitive problems such as in games~\cite{ind12,samo13}, or cooperative tasks that require the coordination of several agents such as in some problems related to evolutionary robotics~\cite{kala12,pan05}, or for situations where the fitness function cannot be designed straightforwardly. Early experimental results~\cite{hill90,ros97} have produced considerable optimism about coevolutionary designs, while more recent works ~\cite{mic09,popo10} were rather to cast some doubt regarding easily understandable (and applicable) coevolutionary problem solvers. The main difficulty appears to be the complex notion of (co--) evolutionary progress and genetically inheritable superiority. Several concepts have been proposed to entangle this complexity and remedy its effects, see the discussion in Sec. \ref{sec:issues}.

A central issue in evolutionary computation is to have a theoretical framework for describing and understanding the evolutionary dynamics underlying evolutionary search. One fundamental way for addressing this issue is the concept of fitness landscapes, which has been successfully applied to gain insight into the evolutionary search processes solving static~\cite{kall01,smit02} and dynamic~\cite{rich08} optimization problems.  Consequently, fitness landscapes have also been proposed to understand coevolutionary processes~\cite{popo04,popo05}, while most recently it has been suggested to employ dynamic landscapes~\cite{rich14}. In this paper, the concept of dynamic landscapes is applied and extended. The aim of this approach is twofold. A first is that dynamic landscapes offer the possibility of studying the dynamics of two major characterizing quantities in coevolutionary algorithms, subjective fitness and objective fitness. A second is that the landscapes obtained enable an analytic treatment valid for all possible individuals of a population (for instance using landscape measures, see e.g.~\cite{mal13} for a recent review). These analytic results may establish a quantification for the differences between the objective landscape describing the problem to be solved and the subjective landscape describing how the coevolutionary algorithm perceives the problem.  

In pursuing these aims,  Sec. \ref{sec:issues} first reviews  some of the issues in coevolution  and highlights the complex and possibly even pathological behavior that can sometimes be observed in coevolutionary runs. Also,  ideas to explain and predict these behavioral features are discussed, namely solution concepts, interactive domains and objective as well as subjective fitness. In Sec. \ref{sec:model} we consider simple  models to be employed in the numerical experiments studying fitness landscapes in coevolution. Such abstract and conveniently experimentable coevolutionary models have been named minimal substrates by Watson and Pollack~\cite{wat01}; this term is adopted here. The fitness landscapes of these models are reported in Sec. \ref{sec:land}. As  the respective landscapes of the interacting species are coupled and dynamically deforming each other, we  refer to such landscapes as codynamic.  In addition, similarity measures of codynamic fitness landscapes are introduced and experimentally studied. The paper ends with discussing  results and drawing conclusions.

\section{Issues in coevolution} \label{sec:issues}
Coevolutionary algorithms (CEAs) differ fundamentally from evolutionary algorithms (EAs) about the way fitness is assigned to individuals. The individuals of an EA may inhabit the search space $S$. For static optimization problems, each of its points $x \in S$ possesses uniquely a fitness value $f(x)$, which is assigned to the individual if the move operators of the algorithm bring the individual to that point in a given generation $k \in \mathbb{N}_0$. For the search space being metric (or otherwise equipped with a neighborhood structure $n(x)$), these  elements cast the (static) fitness landscape $\Lambda_s=(S,f(x),n(x))$.  Dynamic optimization problems deviate from the static view by the fitness of an individual  depending on time, which can be linked to generational time $k$, that is $f(x,k)$. Such a dynamic landscape still consists of search space, fitness function and neighborhood structure, but additionally includes generational time and rules for changing fitness with time. Anyway, fitness is always objective in that the search space point (and possibly generational time) alone defines it. In other words, fitness is a property of a search space point, every individual has the same fitness value if it is situated at the same search space point (and as long as fitness does not change dynamically), and the fitness of one individual does not depend on the fitness of other individuals. 

Contrary to the objective fitness of EAs,  CEAs assign fitness that is subjective.  In coevolution the fitness of an individual is obtained with respect to the fitness (and possibly the search space points) of other individuals. These other individuals,  which are called evaluators,  do usually not belong to the same population as the individual for which fitness is to be evaluated, but to a coevolving population. As a consequence, the fitness of an individual at a given generation  depends on which evaluators are taken, and on the fitness these evaluators have. As the individuals that form the pool of possible evaluators undergo evolutionary development themselves, the fitness value of a search space point (and hence of the individual situated at this point) may vary with the selection, which makes the fitness subjective. For describing the process of obtaining subjective fitness, the framework of 
 interactive domains and 
solution concepts has been proposed~\cite{popo10}. This framework  replaces the fitness function and sets out the rules for assigning fitness values to individuals.    
The interactive domain
defines how the reciprocal actions between
individuals of one population with evaluators from the other are
organized and how the solution of the interaction is calculated. The solution concepts formalizes
how the solution translates to (personal or collective) fitness of
the individuals, how these fitnesses can be compared and interpreted over the entire coevolutionary run and whether or not the comparison indicates coevolutionary progress.   
To establish coevolutionary progress, however, is sometimes difficult. Solving a (maximization) problem  means finding the search space points with highest fitness -- the peaks in the fitness landscape.  The objective fitness of EAs allows  deducing (evolutionary) progress by a simple comparison of the fitness values -- the higher the value the more likely a peak is detected. Also CEAs aim at finding individuals with highest objective fitness. However, the subjective fitness used to drive the CEA is the result of specific interactions with other coevolving individuals. It may hence be  incomplete and inconsistent with respect to the objective fitness, which is obtainable, at least in principle, by the combination of all possible interactions. Therefore, numerical experiments with CEAs sometimes show pathological features of behavior devoid of stable progress, for instance
 cycling, overspecialization and disengagement~\cite{mic09,popo10,wat01}. All these coevolutionary failures are a direct consequence of the uncertainty connected to the question whether progress in subjective fitness also implies progress in objective fitness.

\section{Simple coevolutionary models} \label{sec:model}
To study essential features of coevolution in numerical experiments requires appropriate models. Particularly for studying the connection between subjective and objective fitness, it is desirable that both quantities can be determined in a fast and easy way. Therefore,  problem description from application domains such as coevolutionary games~\cite{ind12,samo13} and robotics~\cite{kala12,pan05} are less  suitable because they may need considerable numerical setup and the objective fitness is difficult (if at all) calculable. Following this line of thought, it is interesting to ask what  minimal structural and behavioral requirements are needed to exhibit complex and relevant coevolutionary dynamics. Such models have been named minimal substrates by  Watson and Pollack~\cite{wat01}.
Accordingly, a coevolutionary minimal substrate is a simple and abstract model of coevolution which defines an interactive domain and a solution concept, exhibits relevant codynamic features and allows experimental studies of the relationships between subjective and objective fitness.  
In the following, and in addition to the initial model~\cite{wat01}, we recall and  interpret other coevolutionary models proposed earlier~\cite{popo04,popo05} as minimal substrates and introduce some modifications  to these models. 

The optimization problems solvable by CEAs can be classified into two groups: 
compositional problems (in which fitness of an individual is assigned  by an interaction that forms a composite  or team) and test--based problems (where the interaction involves a challenge or test)~\cite{popo10}. Next, simple models for both groups of coevolutionary problems are discussed.
For the group of test--based problems, we consider number games~\cite{dejong07,wat01}. The game studied here has two populations $P_1$ and $P_2$ that inhabit the search spaces $S_x$  and $S_y$, respectively. Both search spaces are one--dimensional and real--valued. At each generation $k=0,1,2,\ldots$,   the individuals
of population $P_1(k)$ can take possible values $x \in S_x$ and
 the population $P_2(k)$ may have values $y \in S_y$. We define identical objective fitness functions over both search spaces, that is $f_{obj}(x)$ over $S_x$ and  $f_{obj}(y)$ over $S_y$, which consequently cast objective fitness landscapes. The subjective fitness for both populations is the result of  an interactive number game. Therefore,  for each calculation of the subjective fitness $f_{sub}(x)$ for an individual from $P_1$, a sample  $s(P_2)$ of individuals from $P_2$ is randomly selected. This sample is statistically independent from the sample for the next calculation. Denote $\mu$ the size of the sample $s(P_2)$ out of $\lambda$ individuals in $P_2$ in total, with $\mu\leq \lambda$. Assigning fitness $f_{sub}(y)$ for the individuals of $P_2$ is likewise but reversed with using statistically independent samples $s(P_1)$ from $P_1$.

 The interactive domain of the number game considered defines that the fitness  $f_{sub}(x)$ with respect to the sample $s(P_2)$ is calculated by counting the (mean) number of members in $s(P_2)$ that have a smaller objective fitness $f_{obj}(s_i(P_2))$, $i=1,2,\ldots,\mu$, than the objective fitness $f_{obj}(x)$: 
\begin{equation}\label{eq:subfitnnumber} f_{sub}(x)=\frac{1}{\mu} \sum_{i=1}^{\mu} \text{score}(x,s_i(P_2)) \end{equation} with $\text{score}(x,s_i)=\left\{ \begin{array}{ll}1 &\text{if} \quad f_{obj}(x)>f_{obj}(s_i) \\ 0 &\text{otherwise}\end{array} \right.$. The  fitness $f_{sub}(y)$ is calculated accordingly from  (\ref{eq:subfitnnumber}) where $y$ and $s_i(P_1)$ replace $x$ and $s_i(P_2)$.  
 The subjective fitness (\ref{eq:subfitnnumber}) has some interesting properties. It is a unitary function $f(x)=\mathbb{R} \rightarrow [0,1]$ for every $f_{obj}$, which eases comparing variants of $f_{sub}$ based on different $f_{obj}$.
The subjective fitness $f_{sub}$ converges to the objective fitness $f_{obj}$ for $f_{obj}$ also being a unitary function,  the sample $s(P_1)$ being large, and the distribution of  $f_{sub}$ over $s_i(P_1)$ matching the distribution of  $f_{obj}$  over $S_x$, where the $x \in S_x$ should be taken to resemble a uniform distribution of $f_{obj}$ on the interval $[0,1]$.

\begin{figure}[thb]
\centering
\includegraphics[trim = 0mm 1mm 0mm 8mm,clip, width=6.0cm, height=3.9cm]{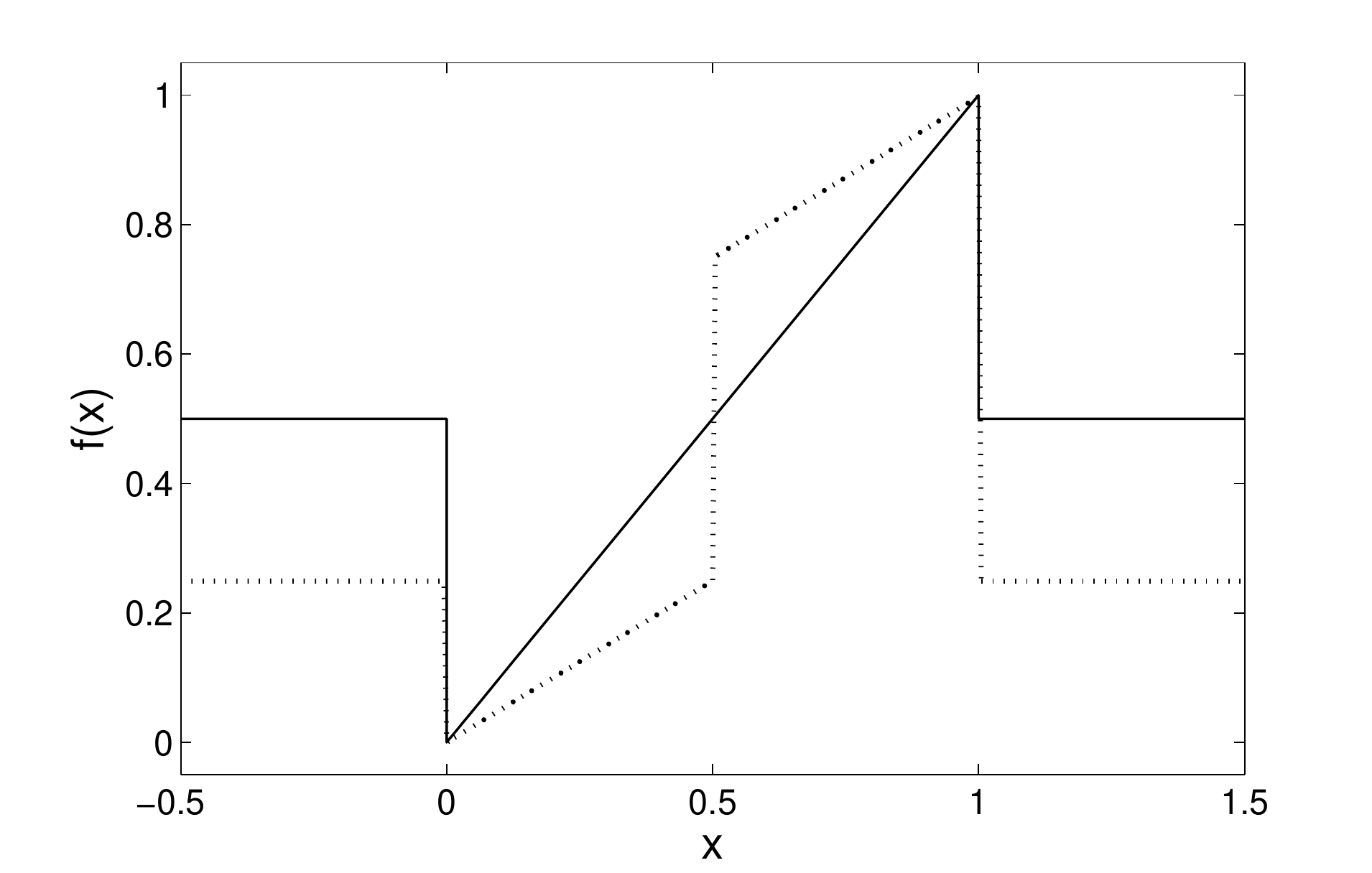} 

(a)

\includegraphics[trim = 0mm 1mm 0mm 8mm,clip, width=6.0cm, height=3.9cm]{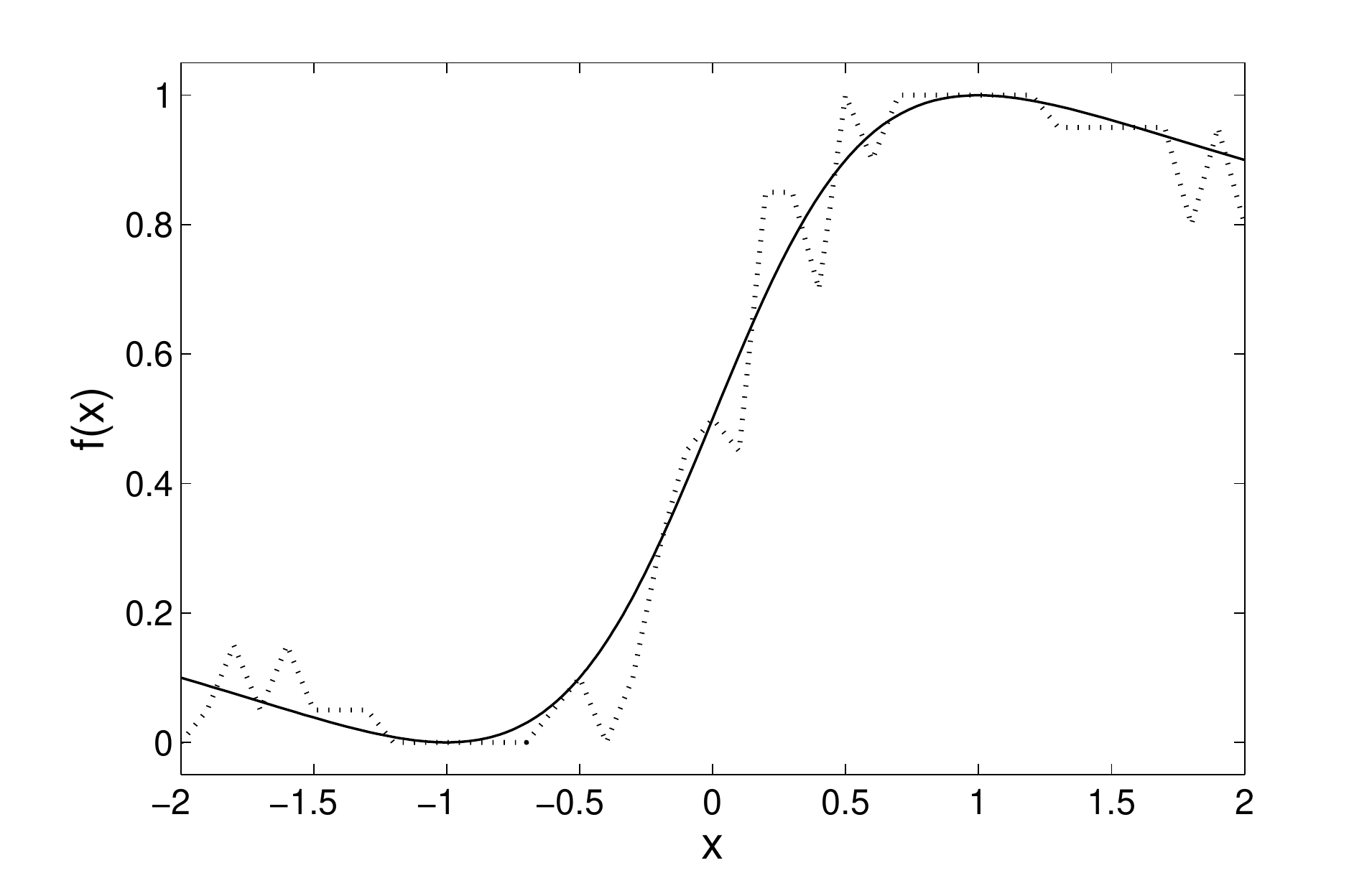} 

(b)
\caption{Objective (solid line) and subjective (dotted line) fitness functions of test--based problems. (a) The piece--wise linear function (\ref{eq:crisp}). (b) The smooth function (\ref{eq:smooth}). }
\label{fig:numb}
\end{figure}

For $f_{obj}(x)=x$, we obtain the number game introduced by Watson and Pollack~\cite{wat01}. The objective fitness function  $f_{obj}(x)=x$ has two optima, one minimum and one maximum. These optima, however, are for the smallest and the largest element in the search space $S_x$, that is, on the boundary of any admissible domain. To numerically obtain these optima in experiments, the locations of the optima require to define (and algorithmically enforce) a bounded search space. This, in turn, ultimately entails a constrained optimization problem and somehow makes the problem setting more complicated than desirable. Therefore, a modification is considered with the piece-wise linear function \begin{equation}f_{obj}(x)=\left\{ \begin{array}{ll} x &\text{for} \quad 0 \leq x\leq 1 \\ 0.5 &\text{otherwise}\end{array} \right..  \label{eq:crisp}\end{equation} This objective fitness has a minimum at $x=0$ with $f_{obj}(0)=0$ and a maximum at $x=1$ with $f_{obj}(1)=1$, and levels off  to a mid--level value of $f_{obj}(x)=0.5$ for $x \rightarrow \pm \infty$.  As a second example of objective fitness the smooth function \begin{equation}
f_{obj}(x)=\frac{1}{2}+\frac{x}{1+x^2} \label{eq:smooth}\end{equation} is taken. It also has two optima, a  minimum at $x=-1$ with $f_{obj}(-1)=0$ and a maximum at $x=1$ with $f_{obj}(1)=1$, and also tends to $f_{obj}(x)=0.5$ for $x$ large in absolute value.  Fig. \ref{fig:numb} shows the objective fitness function (the solid line in the graph) for the test--based coevolutionary problems considered. Whereas the subjective fitness (\ref{eq:subfitnnumber}) may converge to the objective fitness for the conditions given above, in a coevolutionary run both quantities will almost certainly be different. This is because the sample $s(P_1)$ is most likely small compared to the population size of $P_1$, and even smaller compared to the amount of samples needed to cover the entire domain of the search space. Fig.    \ref{fig:numb} gives a realization of the subjective fitness (the dotted line in the graph). This realization is obtained by drawing a medium size sample ($\mu=100$) from a given population ($\lambda=400$) which is uniformly distributed on $S_x$. It can be seen that the subjective fitness resembles the objective fitness. 

For the group of compositional problems, the search spaces $S_x$ and $S_y$ of the coevolving populations may be 
 combined  into one shared landscape $S=\{S_x,S_y\}$.  This might
result in a unique (static) objective landscape for simple coevolutionary
scenarios. The compositional examples considered in this paper also work with coevolving
populations that are one--dimensional and real--valued. Therefore, combining the two
one--dimensional  landscapes leads to a shared two--dimensional objective
landscape.  This approach can be found in previous research~\cite{popo04,popo05} on
understanding coevolutionary phenomena by fitness
landscapes. We interpret these examples as compositional  minimal substrates and employ 
ridge functions as suggested in~\cite{popo04,popo05}.
\begin{figure}[thb]
\centering
\includegraphics[trim = 0mm 1mm 0mm 8mm,clip, width=6.0cm, height=3.9cm]{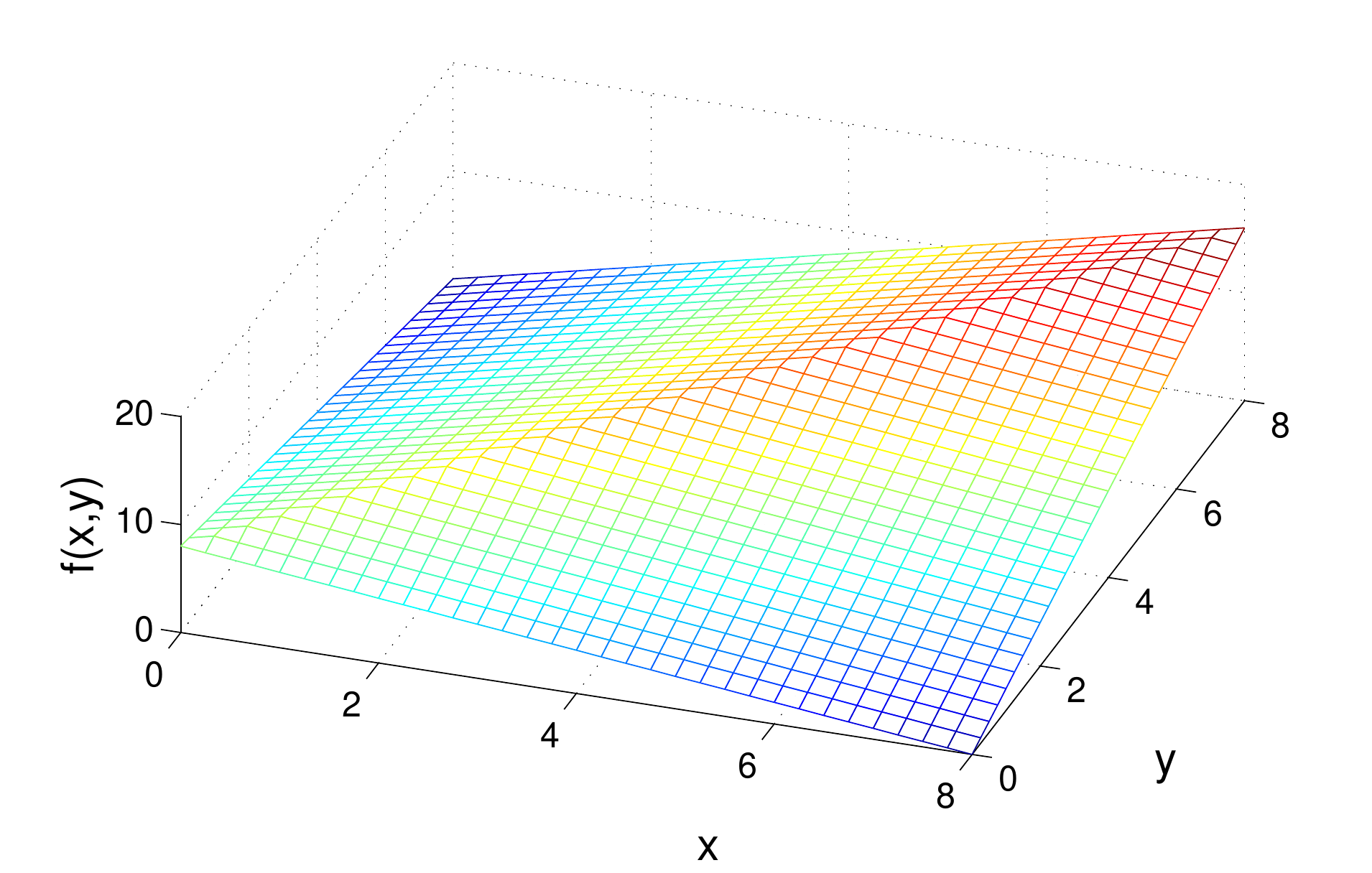} 

(a)

\includegraphics[trim = 0mm 1mm 0mm 8mm,clip, width=6.0cm, height=3.9cm]{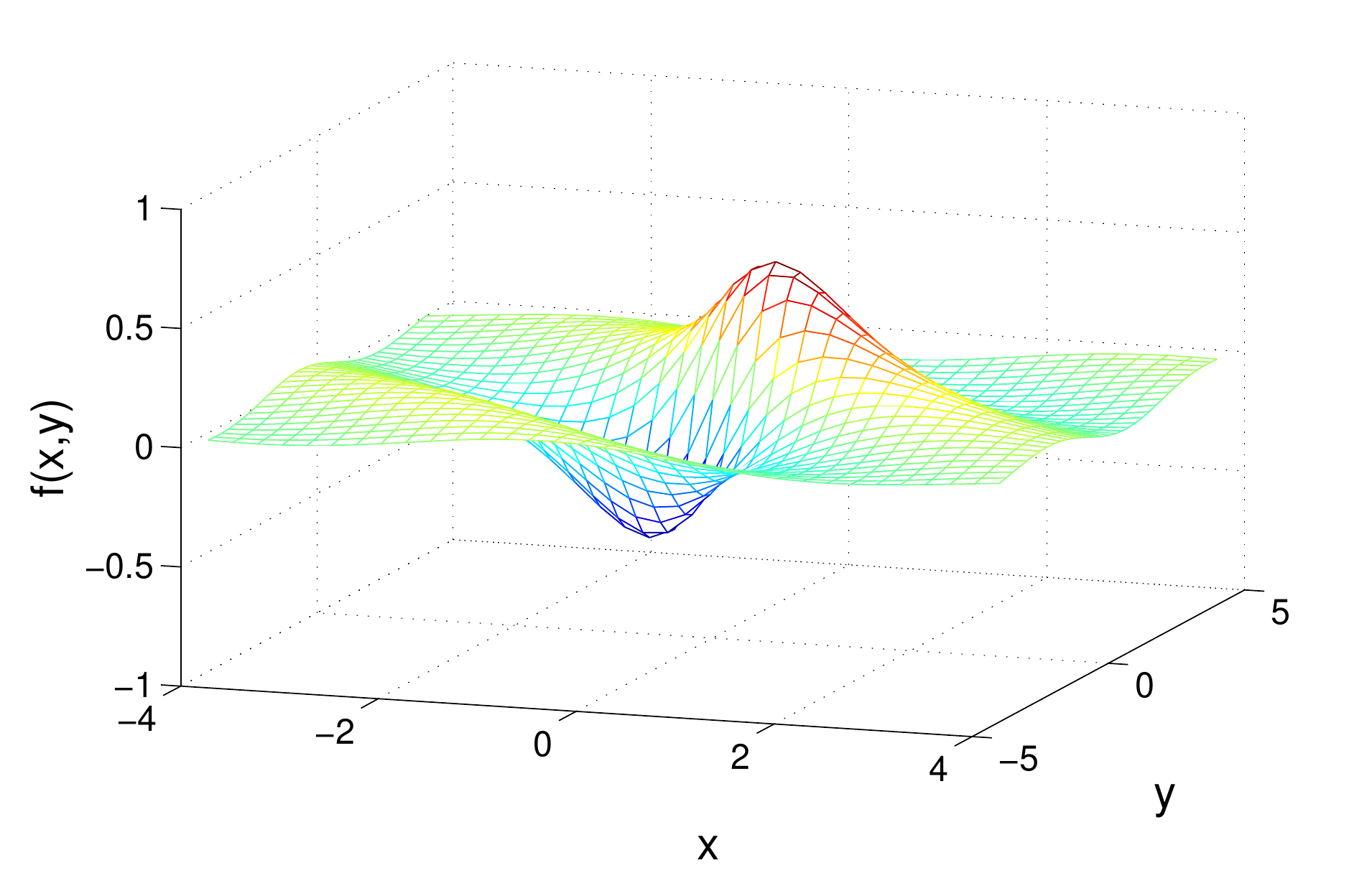} 

(b)
\caption{Shared objective fitness functions of compositional problems. (a) The ridge function  (\ref{eq:ridge}) for $n=8$. (b) The sinusoid function  (\ref{eq:sin}). }
\label{fig:ridge}
\end{figure}
 The simplest function has one ridge: \begin{equation} f_{obj}(x,y)=\left\{ \begin{array} {lcc}n+2 \min{(x,y)}-\max{(x,y)}  \\ \qquad \qquad \text{for} \quad 0 \leq (x,y)\leq n \\ n \quad \text{otherwise}\end{array} \right. ,\label{eq:ridge} \end{equation}
with $x,y \in \mathbb{R}$ and $n$ is a parameter that sets the size and the hight of the landscape (see Fig. \ref{fig:ridge}a). The landscape has a single maximum at $f_{obj}(n,n)=2n$ and a ridge diagonally from $f_{obj}(0,0)=n$ to $f_{obj}(n,n)$ that separates two planar surfaces. There are two minima at $f_{obj}(0,n)=f_{obj}(n,0)=0$.  Outside the square $0 \leq (x,y)\leq n$, the landscape has the mid--level value $f_{obj}(x,y)=n$ that ensures that the optima do not lie on the boundary of the admissible domain.  Equation (\ref{eq:ridge}) is the fitness function for both populations $P_1$ and $P_2$ and can be interpreted as the static shared objective fitness landscape defined over $S=\{S_x,S_y\}$. As a second example the smooth shared objective landscape
\begin{equation} f_{obj}(x,y)=\frac{\sin(x+y)}{1+x^2+y^2} \label{eq:sin}\end{equation}
is analyzed; see Fig. \ref{fig:ridge}b. It has a global minimum at $f_{obj}(-0.4925,-0.4925)=-0.5611$, a global maximum at $f_{obj}(0.4925,0.4925)=0.5611$, and levels off to $f_{obj}(x,y)=0$ for $(x,y)$ large in absolute value. 

The subjective fitness for each population is calculated by using the shared objective fitness functions  (\ref{eq:ridge}) or (\ref{eq:sin}) and inserting a value obtained by a metric on the population of the respective other population instead of the required second variable. Thus, for calculating the subjective fitness of $P_1$, a metric  $m(P_2)$ on $P_2$ is taken: 
\begin{equation} f_{sub}(x)=f_{obj}(x,m(P_2)). \label{eq:subfitridge}\end{equation}
Replacing $m(P_1)$ and $y$ for $x$ and $m(P_2)$ in (\ref{eq:subfitridge}) yields the subjective fitness  $f_{sub}(y)$. 
The metric $m(P_2)$ used in the numerical experiments reported here is to identify and employ the individual with maximal or minimal fitness at a given generation. In some sense this means that all individuals of the other population act as evaluators by rating and presenting its best member.

Another significant issue in coevolutionary scenarios is the character of the interaction.  A main classification  is cooperative or 
competitive interaction~\cite{chen13,popo10,shi12}. Cooperative means that the
individual and the evaluators interact and collaborate to solve a
problem that is harder or impossible to solve by each of them
alone. The better they support each other and perform together, the higher the reward and
hence the fitness. In other words, both populations work towards the same aim.  In competitive
interaction the individual is rewarded for out--performing the
evaluators, which sometimes means that the fitness of one
individual is increased at the expense of the others.   The main feature here is that the aims of the populations involved are conflicting.

Interestingly, for the simple examples of coevolutionary models considered here, either cooperative
or competitive   interactions can be imposed  in an abstract way. For the compositional minimal substrates given by   (\ref{eq:subfitridge}), the question of cooperative or competitive interaction can be decided by the properties of the metrics $m(P_1)$ and $m(P_2)$. 
These metrics answer which  individual of either population is the best,
$x_{best}(k)=m(P_1(k))$ and $y_{best}(k)=m(P_2(k))$, respectively. As shown
in~\cite{popo04,popo05}, a cooperative interaction is imposed if
the task for both populations is  the same, that is, both are to
find the maximum or minimum of the objective fitness function (\ref{eq:ridge}) and (\ref{eq:sin}).
A competitive interaction takes place if one population is to
search for the maximum of (\ref{eq:ridge}) and (\ref{eq:sin}), while the other is to
find the minimum of (\ref{eq:ridge}) and (\ref{eq:sin}). In a similar manner, 
for the test--based minimal substrates given by (\ref{eq:subfitnnumber}), a cooperative interaction can be observed if both populations mean to find the maximum (or minimum) of the objective fitness. A competitive scenario occurs if one population searches for the maximum while the other intends to find the minimum. 

So far, the interactive domains and the character of interaction of the minimal substrates were laid out.
As evaluation and subsequent updating of fitness in one population requires evaluators from the other population, the question of timely order and sequence becomes an issue. 
The models we consider here have a synchronous mode of evolutionary flow. We consider synchronization that takes place after a generation of both populations (shared synchronization). Such synchronization points mean that both populations evolve along the conventional EA's generational process (fitness evaluation followed by selection, possibly recombination and  mutation) and communicate via delivering evaluators to the respective other population. This implies that the fitness of $P_1(k+1)$ and $P_2(k+1)$ is calculated using evaluators from $P_2(k)$ and $P_1(k)$, respectively.  As the populations take turns in evolving, this creates a coupling via the (time--dependent) fitness values of the respective population. As an effect, both populations coevolve, and the landscapes show codynamics. 

How this codynamics is reflected in the subjective fitness landscapes is analyzed using numerical experiments with a CEA. 
  The experimental results reported are obtained for an algorithm with separated populations that undergo selection and mutation independent from each other. The coevolutionary interactions are carried out as described above. Unless otherwise stated the population size is $\lambda_1=\lambda_2=24$ and for the test--based minimal substrate there are $\mu_1=\mu_2=12$ evaluators. Selection is tournament with size $2$  and mutation is Gaussian with mutation probability $0.5$ and mutation strength $0.1$. In accordance to other studies~\cite{wat01}, no recombination is used.  

\section{Codynamic fitness landscapes} \label{sec:land}
\subsection{Pictorial results}
For the test--based problem, the subjective fitness  of population $P_1(k+1)$ at generation $k+1$ is calculated according to (\ref{eq:subfitnnumber}) using a sample $s(P_2(k))$ from the population $P_2(k)$ and yields the landscape: 
\begin{equation}\label{eq:subfitnnumber1} f_{sub}(x,k+1)=\frac{1}{\mu} \sum_{i=1}^{\mu} \text{score}(x,s_i(P_2(k))). \end{equation}
The samples $s(P_2(k))$ are statistically independent over the individuals for which fitness is to be assigned. This  means for each individual in every generation (and every coevolutionary run), there is a specific (subjective) fitness landscape. Stated like that it seems hopeless to draw any useful information from analyzing such landscapes. However, while the samples are statistically independent, the possible members drawn and used as evaluators are not as they belong to the coevolving population. This implies that the subjective fitness landscape may follow certain patterns, and that these patterns reveal the general topology of the (subjective) landscape, at least as the result of averaging or another analyzing method.  
The subjective fitness of $P_2(k+1)$ is  calculated likewise by  (\ref{eq:subfitnnumber1}), but by using a sample   $s(P_1(k))$. 
Note that the character of the interaction (cooperative or competitive) may influence the average composition of the population and consequently the (average) selection of evaluators. For instance, if in competitive interaction the population $P_1$ searches the minimum and $P_2$ the maximum, then the evaluators drawn from $P_2$ will on average be larger and generally may have other statistical properties as if  both populations were to find the minimum. This also affects the subjective landscape   (\ref{eq:subfitnnumber1}), albeit in an implicit way only.   Further note that the subjective landscape of $P_1$ and the landscape of $P_2$ are coupled via the evaluators from the respective other population which implies that 
 codynamics occurs between these landscapes. 

For the compositional problem we get for the objective fitness function (\ref{eq:ridge})
the subjective fitness landscape of population $P_1$  
\begin{equation} f_{sub}(x,k+1)=n+2 \min{(x,y_{best}(k))}-\max{(x,y_{best}(k))}, \label{eq:ridge1} \end{equation} while for population $P_2$ we obtain
$f_{sub}(y,k+1)$ and replace $x$ and $y_{best}(k)$ by $y$ and $x_{best}(k)$ in (\ref{eq:ridge1}).
From the perspective of the populations alone it appears that fitness is
calculated on--the--fly while the CEA is
running.  
For the objective fitness function (\ref{eq:sin}), the landscapes read accordingly. 
Also these landscapes are coupled and codynamic. In difference to the test--based landscapes, we have the same landscape for all individuals of each population, but the landscapes still vary over generations and coevolutionary runs.    Furthermore, as the landscapes depend explicitly on $x_{best}$ and  $y_{best}$, cooperative and competitive interaction may explicitly yield  landscapes with different shapes.

\begin{figure}[thb]
\centering
\includegraphics[trim = 0mm 1mm 0mm 8mm,clip, width=6.0cm, height=3.9cm]{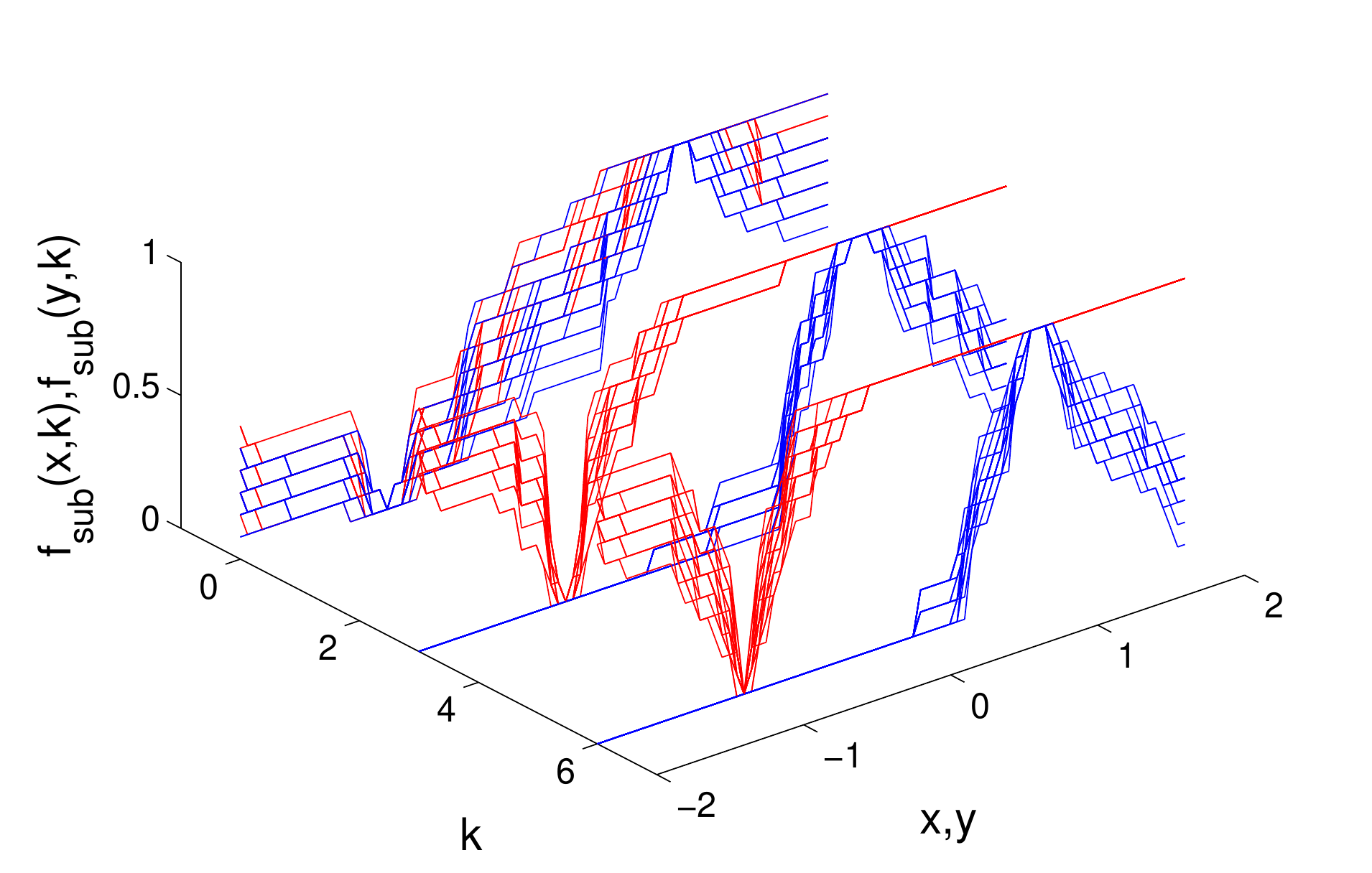} 

(a)

\includegraphics[trim = 0mm 1mm 0mm 8mm,clip, width=6.0cm, height=3.9cm]{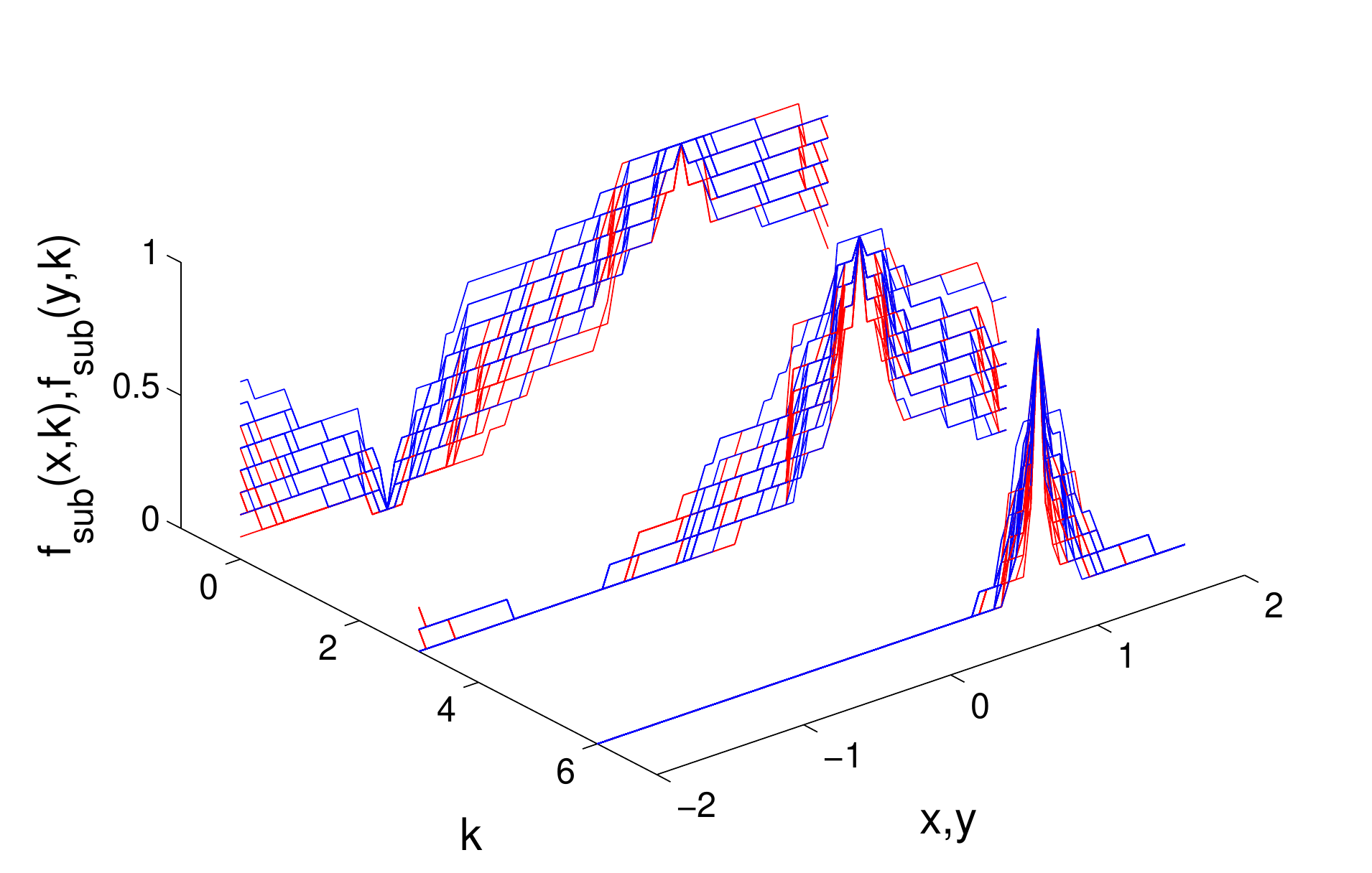} 

(b)
\caption{Realizations of codynamic fitness landscapes of the test--based problem specified by the smooth objective fitness function  (\ref{eq:smooth}). (a) Competitive interaction. (b) Cooperative interaction.}
\label{fig:landcompcoop}
\end{figure}
\begin{figure}[thb]
\centering
\includegraphics[trim = 0mm 1mm 0mm 8mm,clip, width=6.0cm, height=3.9cm]{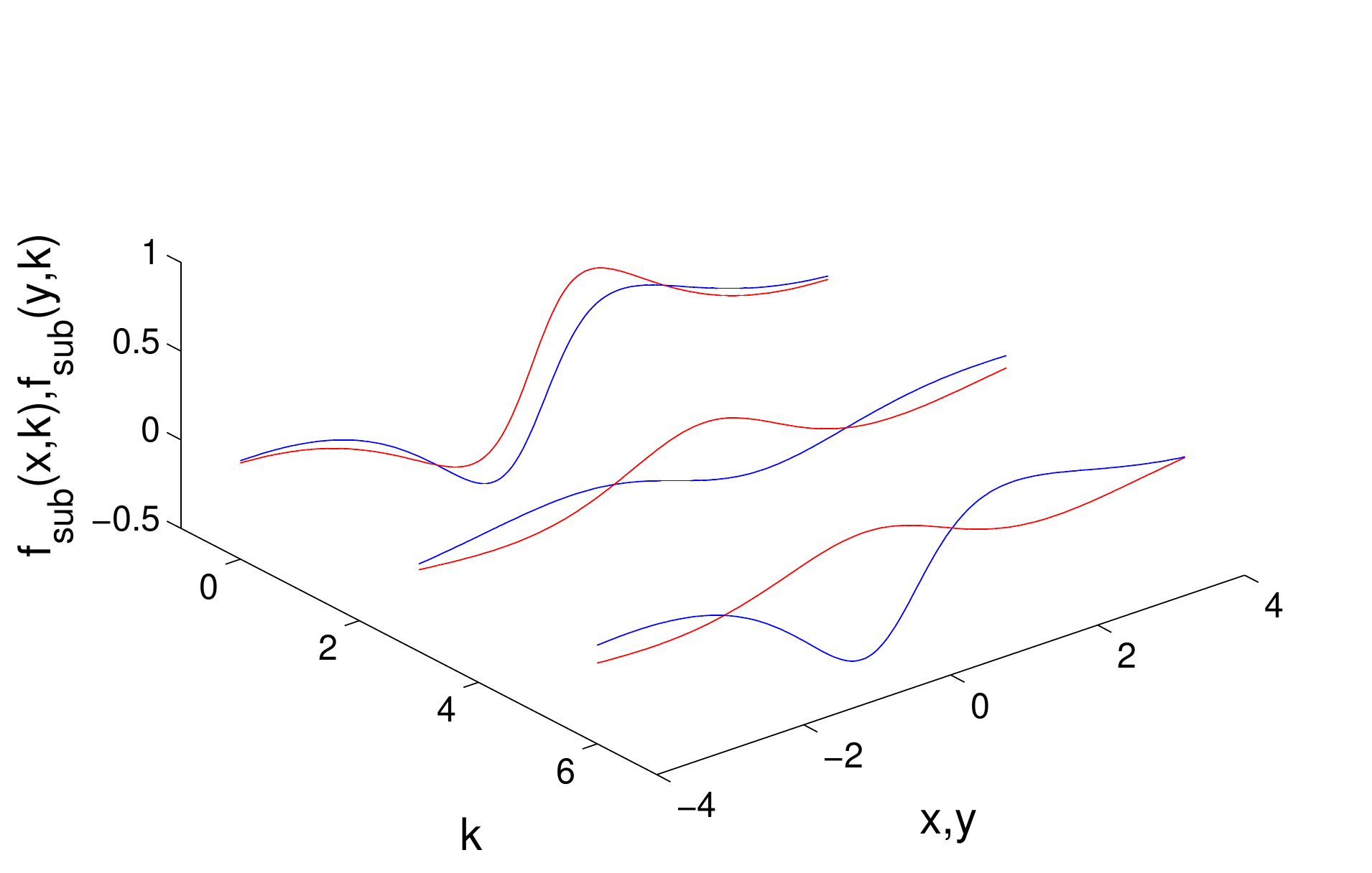} 

(a)

\includegraphics[trim = 0mm 1mm 0mm 8mm,clip, width=6.0cm, height=3.9cm]{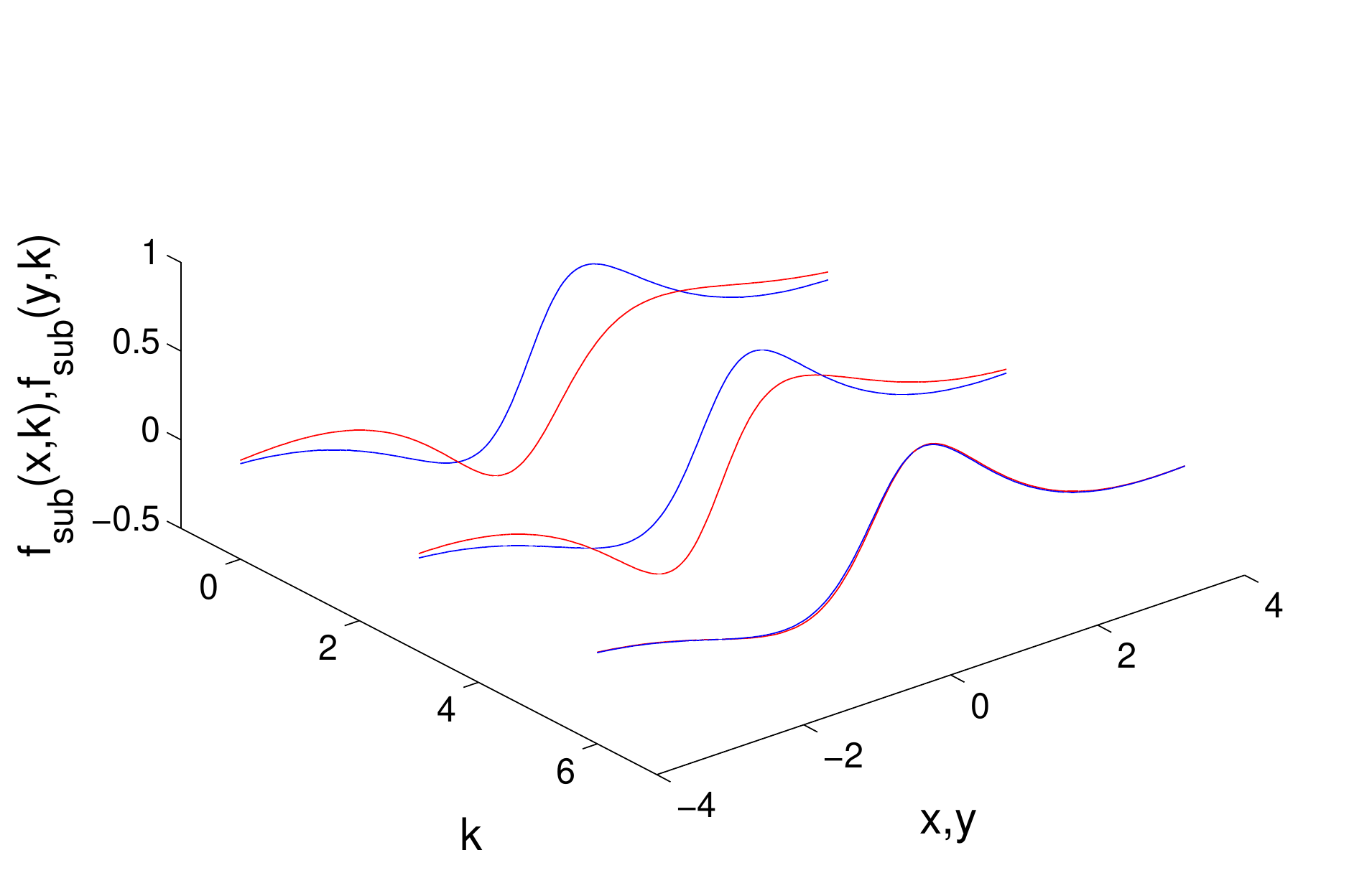} 

(b)
\caption{Realizations of codynamic fitness landscapes of the compositional problem specified by the shared objective fitness function  (\ref{eq:sin}). (a) Competitive interaction. (b) Cooperative interaction.}
\label{fig:landcompcoop1}
\end{figure}

Due to the simplicity of
the examples, the codynamic fitness landscapes can be depicted
as a function of coevolutionary run--time.
Fig. \ref{fig:landcompcoop} shows realizations of the codynamic landscapes of the test--based problem specified by the smooth objective fitness function  (\ref{eq:smooth}) and in Fig. \ref{fig:landcompcoop1} the codynamic landscapes of  the compositional problem specified by the shared objective fitness function  (\ref{eq:sin}) can be seen.
As an illustration of codynamics, both figures show the subjective landscapes $f_{sub}(x,k)$ of population $P_1$ as red lines and   $f_{sub}(y,k)$ of population $P_2$ as blue lines. The landscapes are given for three points in coevolutionary run--time, $k=0,3,6$. They are realizations of codynamic fitness landscapes because they are the result of a single coevolutionary run.  Another run with another initial population may produce slightly different curves. The numerical experiments, however, have shown that certain pattern are preserved over the runs. 

As illustrated in Fig.  \ref{fig:landcompcoop},  we obtain an ensemble of landscapes for each point in time for coevolutionary test--based problems. This ensemble is built by the possibly different landscapes for each sample $s(P_1)$ or $s(P_2)$. Thus, at the utmost there are as many landscapes as individuals in the population, which is $\lambda=24$ for the example.  
However, the scoring function (\ref{eq:subfitnnumber}) that renders subjective fitness from objective fitness implies a discretization, which means that only a finite number of different landscape shapes are possible.  The effect of discretization is clearly visible in Fig. \ref{fig:landcompcoop}. Discretization also  contributes to the deviation between the curves of the subjective fitness landscape and the objective landscape.  The subjective landscape frequently overestimates or  underestimates the objective landscape (compare the curves in Fig. \ref{fig:landcompcoop} with the curve in Fig. \ref{fig:numb}b), which goes along with coevolutionary intransitivity.  Another interesting feature of codynamic landscapes can be seen for coevolutionary run--time going by. 
For the initial population, which most likely has a large diversity, the shape of the subjective landscapes still somehow resembles the shape of the objective landscape, compare to Fig. \ref{fig:numb}b. For time going on the shape of the subjective landscape changes dramatically. In Fig.  \ref{fig:landcompcoop}a competitive interaction is shown where population $P_1$ searches for the minimum and $P_2$ intends to find the maximum. It is visible that the landscape $f_{sub}(x,k)$ contracts around the solution peak, while the landscape  $f_{sub}(y,k)$ does the same around the solution valley. Other topological features of the landscape (for instance the respective valley and peak) are blanked out. It appears as if the coevolutionary process creates the fitness landscape in which the algorithm performs the search. Interestingly, this blanking out effect is also noticeable for cooperative interaction,  see Fig.  \ref{fig:landcompcoop}b. However, here both subjective landscapes evolve along similar pattern. Comparing the figures also reveals that the degree of contraction varies from run to run, which emphasize that each subjective landscape in itself is a realization.  
The subjective landscapes of the compositional problem specified by the shared objective fitness function  (\ref{eq:sin}) are given in Fig. \ref{fig:landcompcoop1} and show a slightly different behavior. Again, the  landscapes $f_{sub}(x,k)$ of population $P_1$  are shown as red lines and   $f_{sub}(y,k)$ of population $P_2$ as blue lines. We have a single landscape for each generation as there is just one landscape for all individuals in compositional coevolution. All subjective landscapes are slices through the shared objective landscape. Thus, 
the landscapes could also be
directly derived from the shared  fitness landscape in Fig.
\ref{fig:ridge}b by looking from the $x$--axis (or $y$--axis) and considering the
value for $y=y_{best}(k)$  (or $x=x_{best}(k)$) as slices of the $S_x$ (or $S_y$) space. Again, a difference in cooperative and competitive coevolution can be observed by the landscapes in cooperation (Fig. \ref{fig:landcompcoop1}b) converging while the landscapes in competition diverging. Also, it might be that the subjective landscape at a given generation does not include the maximum (or minimum) of the objective landscape, thus making it impossible to search for it. 
The codynamic landscapes for the other problems show similar characteristics, but are not depicted here for sake of brevity. 

\subsection{Similarity measures}

As instructive as these pictorial descriptions of landscapes might be, they also clearly show the limitations of geometrical conceptualization.  The pictures are widely open to interpretation and bound to maximally two--dimensional search spaces. Therefore, we next study
properties of codynamic landscapes based on analytic quantities. In doing so we define landscape measures of codynamic landscapes. 
It has been argued that coevolutionary failure, intransitivity  and pathological behavior is a consequence of subjective fitness dissociating from objective fitness~\cite{dejong07,vanwijn08}. Hence, it appears to be interesting to  analyze how measures of similarity between the objective and subjective landscape behave over coevolutionary run--time. As the minimal substrates  allow to analytically describe both the subjective and objective landscape, a calculable (geometric) similarity measure is Euclidean distance ($\text{dist}$), which we define as \begin{equation} \text{dist}(k)=\frac{1}{\text{dist}_{max}}\: \| f_{obj}(x_j)-f_{sub}(x_j,k)\|, \label{eq:dist} \end{equation} where  $x_j$ are a countable number of search space points in $S_x$, and $\text{dist}_{max}$ is the maximal fitness difference in the landscape. For the compositional landscape $f_{obj}(x,y)$, the component $y$ is set to the global $y_{max}$ or $y_{min}$, respectively. For test--based problems, the subjective landscape  $f_{sub}(x_j,k)$ is built by averaging over the samples. We further test two statistical 
difference measures. A first  is Kullback--Leibler divergence ($\text{kld}$), e.g.,
see~\cite{cov06}, p.~19:
\begin{equation}
\text{kld}(k)= \sum_{j} \bar{f}_{obj}(x_j) \: \log_2
\left( \frac{\bar{f}_{obj}(x_j)}{\bar{f}_{sub}(x_j,k) } \right),
\label{eq:kld} 
\end{equation}
which is calculated with a countable number of search space points $x_j$ and normalized subjective and objective fitness values $\bar{f}_{obj}$ and $\bar{f}_{sub}(x_j,k)$.  This normalization allows to view subjective and objective fitness as quantities similar to distributions. Hence, the  $\text{kld}$ in  (\ref{eq:kld}) measures  the entropic distance from  objective fitness to subjective fitness.  As a third similarity measure of landscapes we consider the  Bhattacharyya coefficient ($\text{bhatt}$)~\cite{com03} which assesses the similarity of two probability distributions. It is obtained by partitioning the objective and subjective fitness landscapes into normalized histograms $h_{obj}(x_j)$ and $h_{sub}(x_j,k)$ with bin centers specified by $x_j$ and calculating 
\begin{equation} \text{bhatt}(k)=\sqrt{1-\sum_{j} h_{obj}(x_j) \cdot h_{sub}(x_j,k)}. \label{eq:bhatt}\end{equation} Hence, the $\text{bhatt}$ in (\ref{eq:bhatt}) measures the amount of overlap between objective and subjective fitness. The equations  (\ref{eq:dist}),   (\ref{eq:kld}) and (\ref{eq:bhatt})  are for calculating the measures of the landscape over $S_x$, For the measures over $S_y$, replace $y$ for $x$.
\begin{figure*}[t]

\includegraphics[trim = 0mm 1mm 0mm 2mm,clip, width=6cm, height=3.9cm]{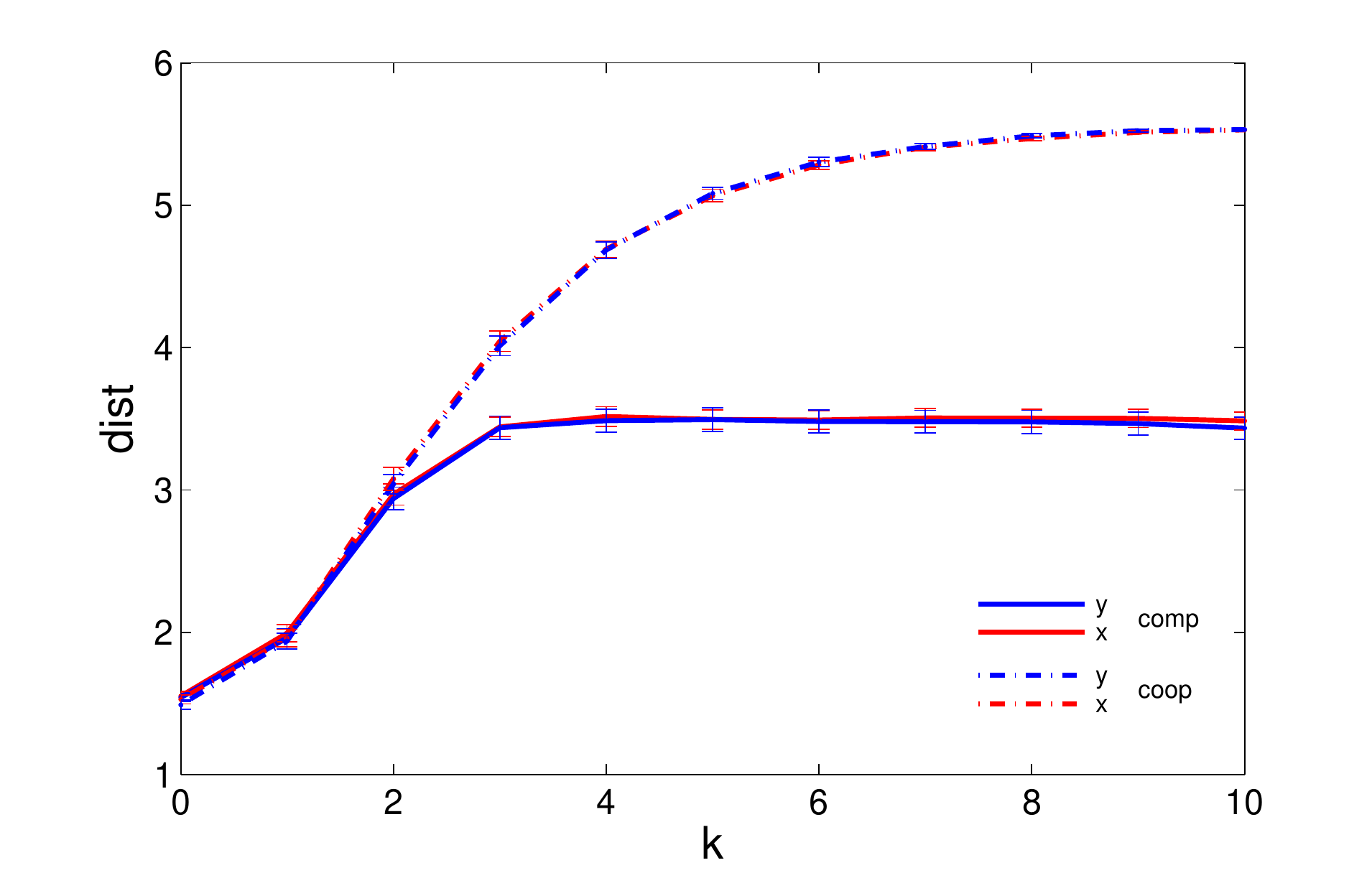} 
\includegraphics[trim = 0mm 1mm 0mm 2mm,clip, width=6cm, height=3.9cm]{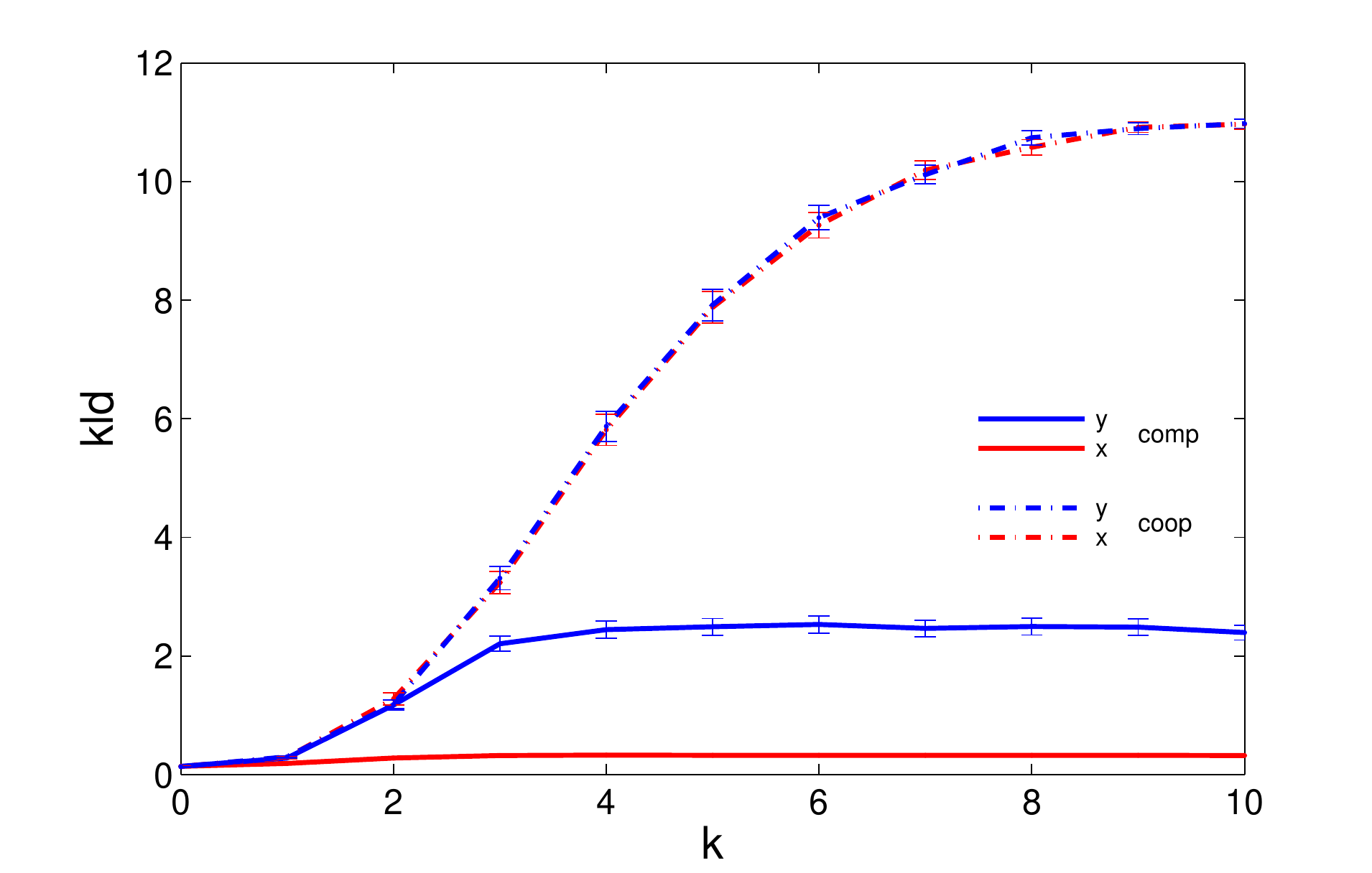} 
\includegraphics[trim = 0mm 1mm 0mm 2mm,clip, width=6cm, height=3.9cm]{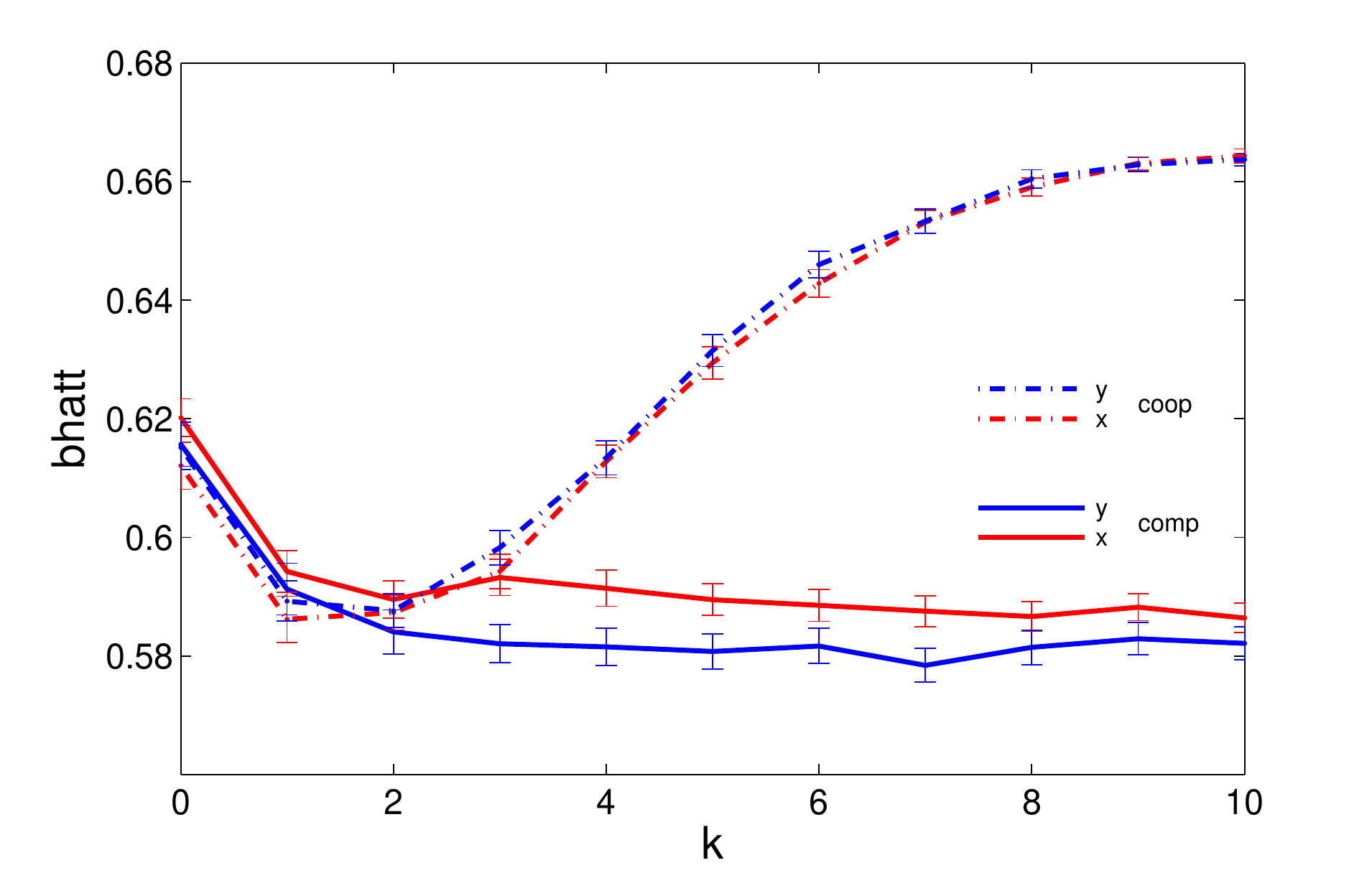} 

\hspace{3cm} (a) \hspace{5cm} (b) \hspace{5cm} (c)

\includegraphics[trim = 0mm 1mm 0mm 2mm,clip, width=6cm, height=3.9cm]{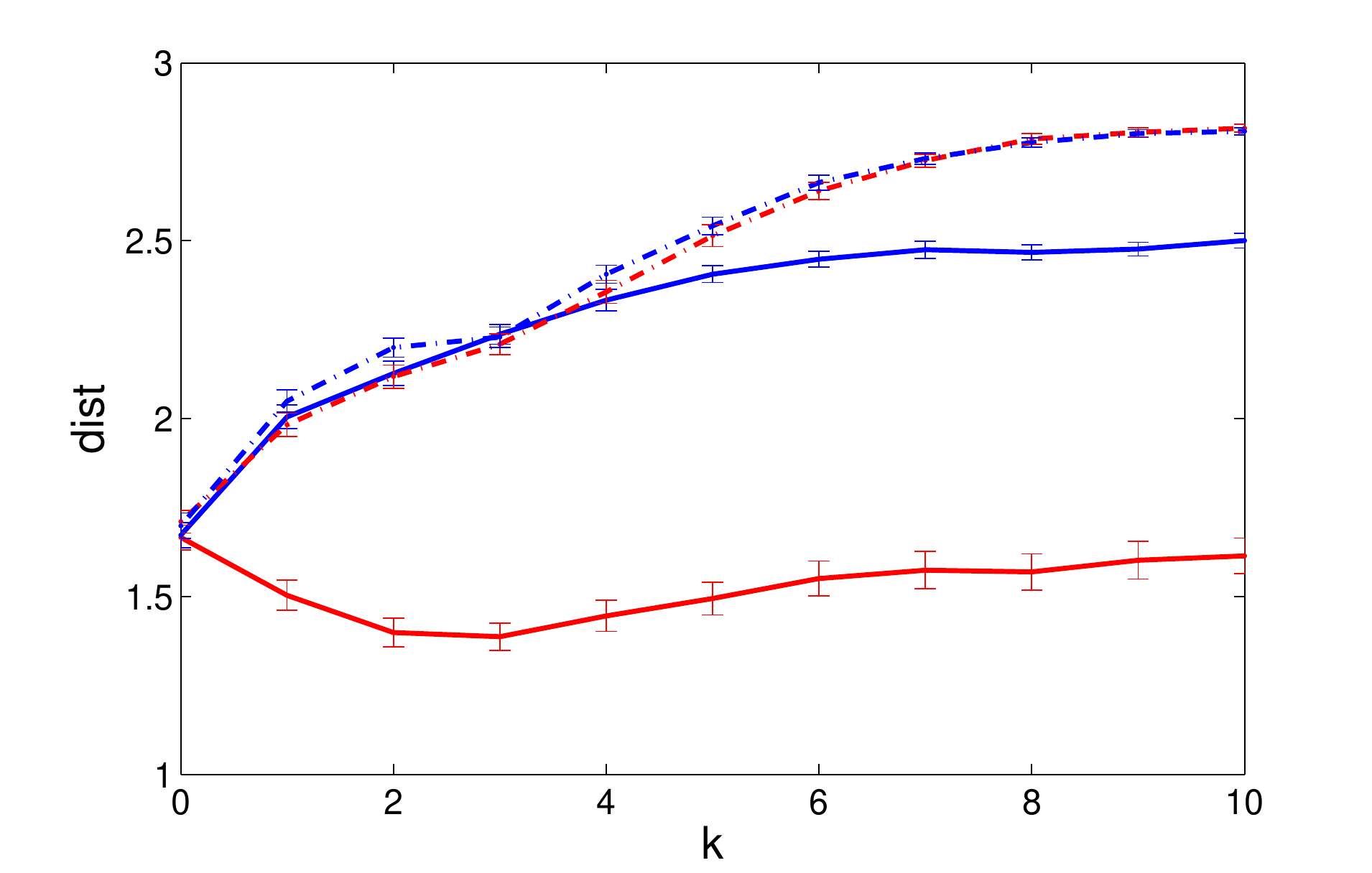} 
\includegraphics[trim = 0mm 1mm 0mm 2mm,clip, width=6cm, height=3.9cm]{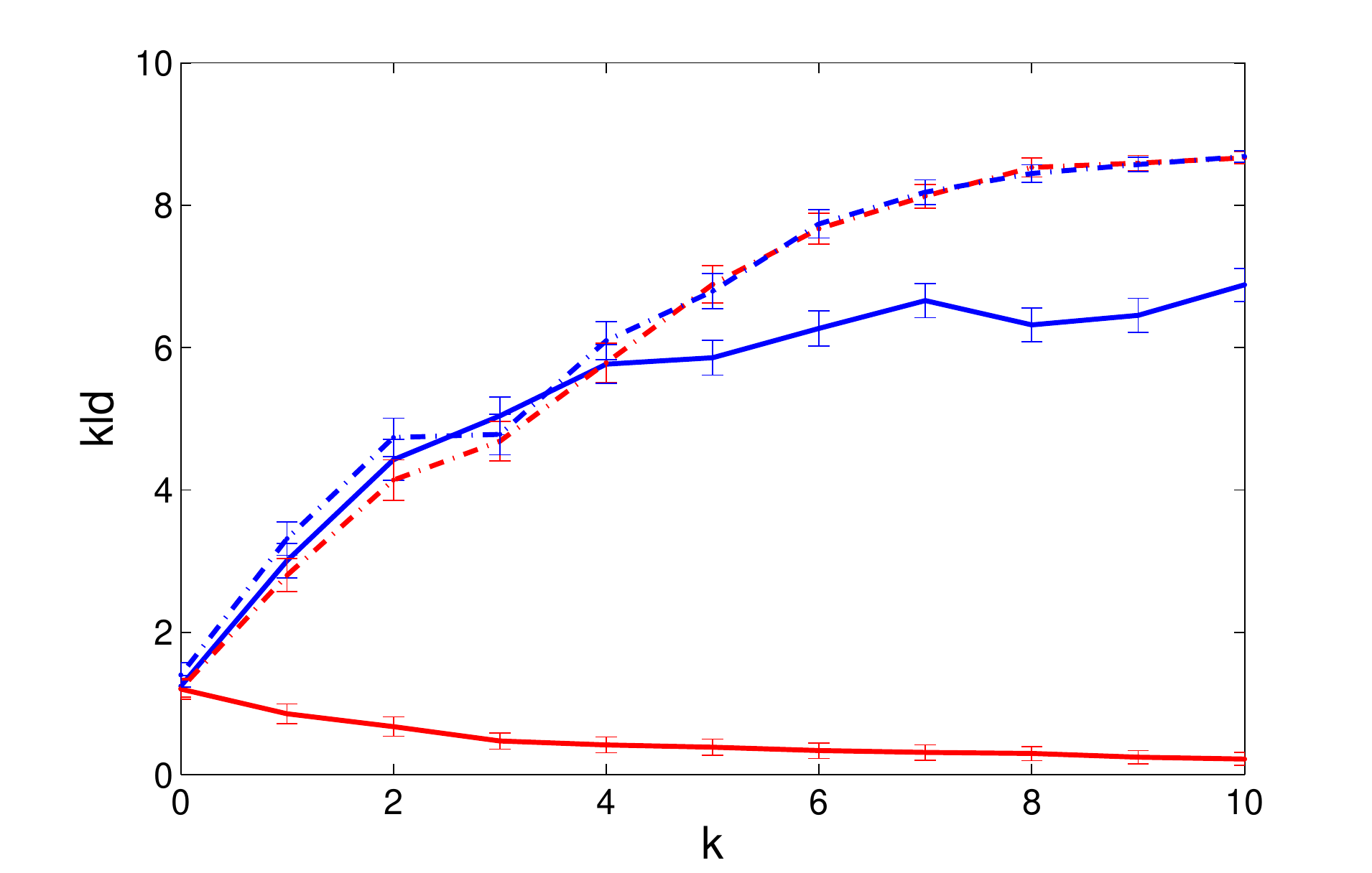} 
\includegraphics[trim = 0mm 1mm 0mm 2mm,clip, width=6cm, height=3.9cm]{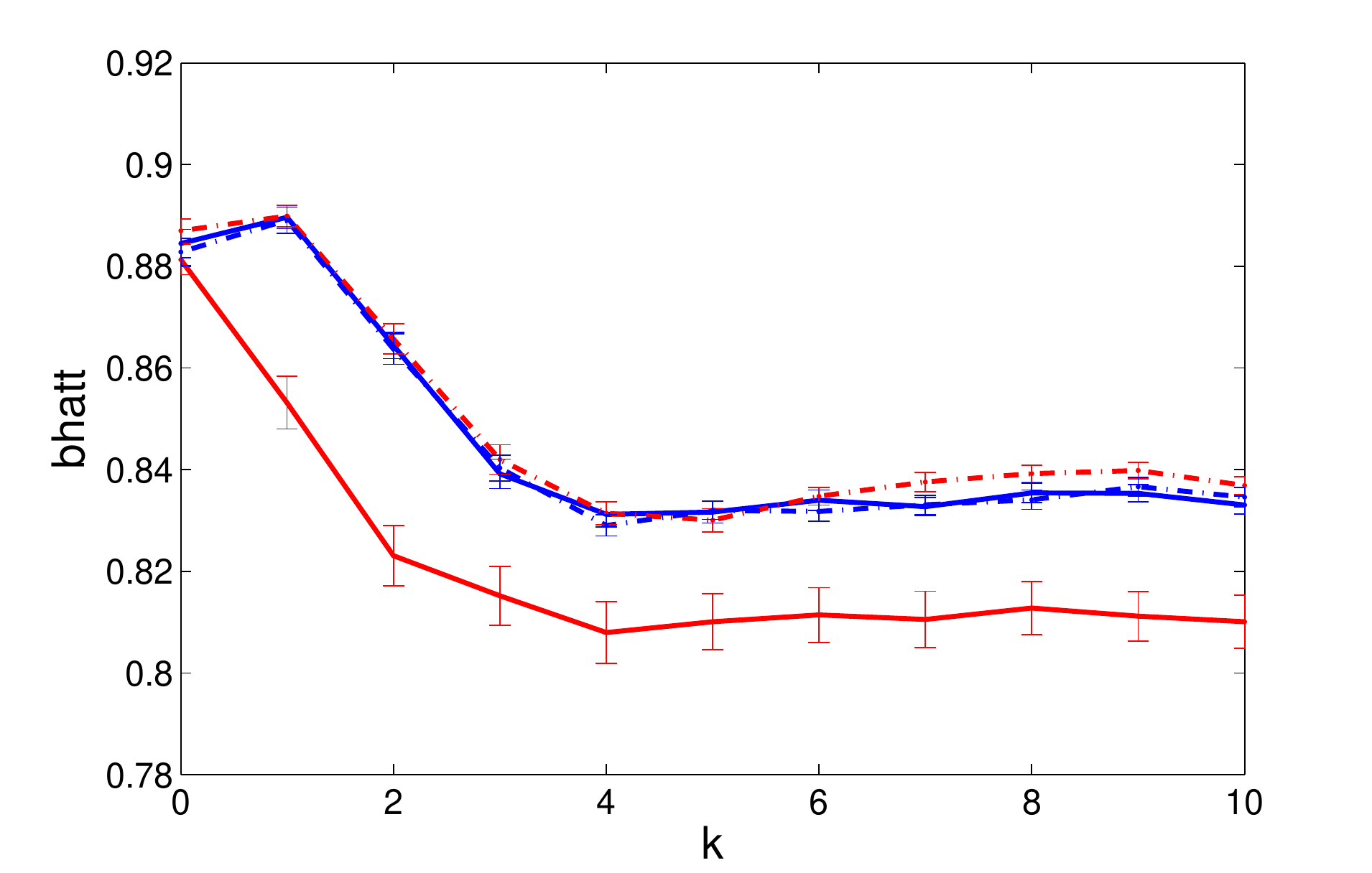} 

\hspace{3cm} (d) \hspace{5cm} (e) \hspace{5cm} (f)

\includegraphics[trim = 0mm 1mm 0mm 2mm,clip, width=6cm, height=3.9cm]{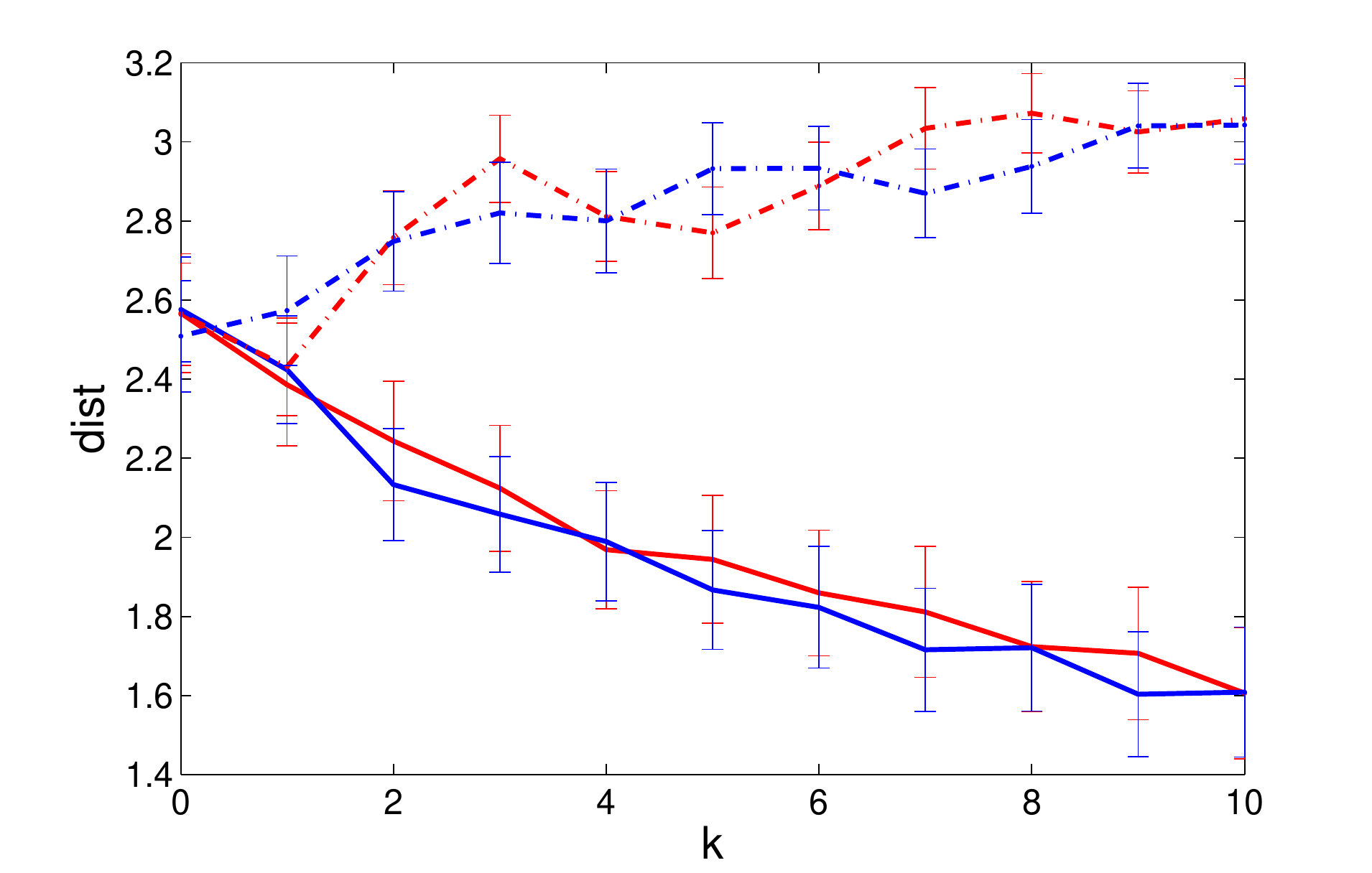} 
\includegraphics[trim = 0mm 1mm 0mm 2mm,clip, width=6cm, height=3.9cm]{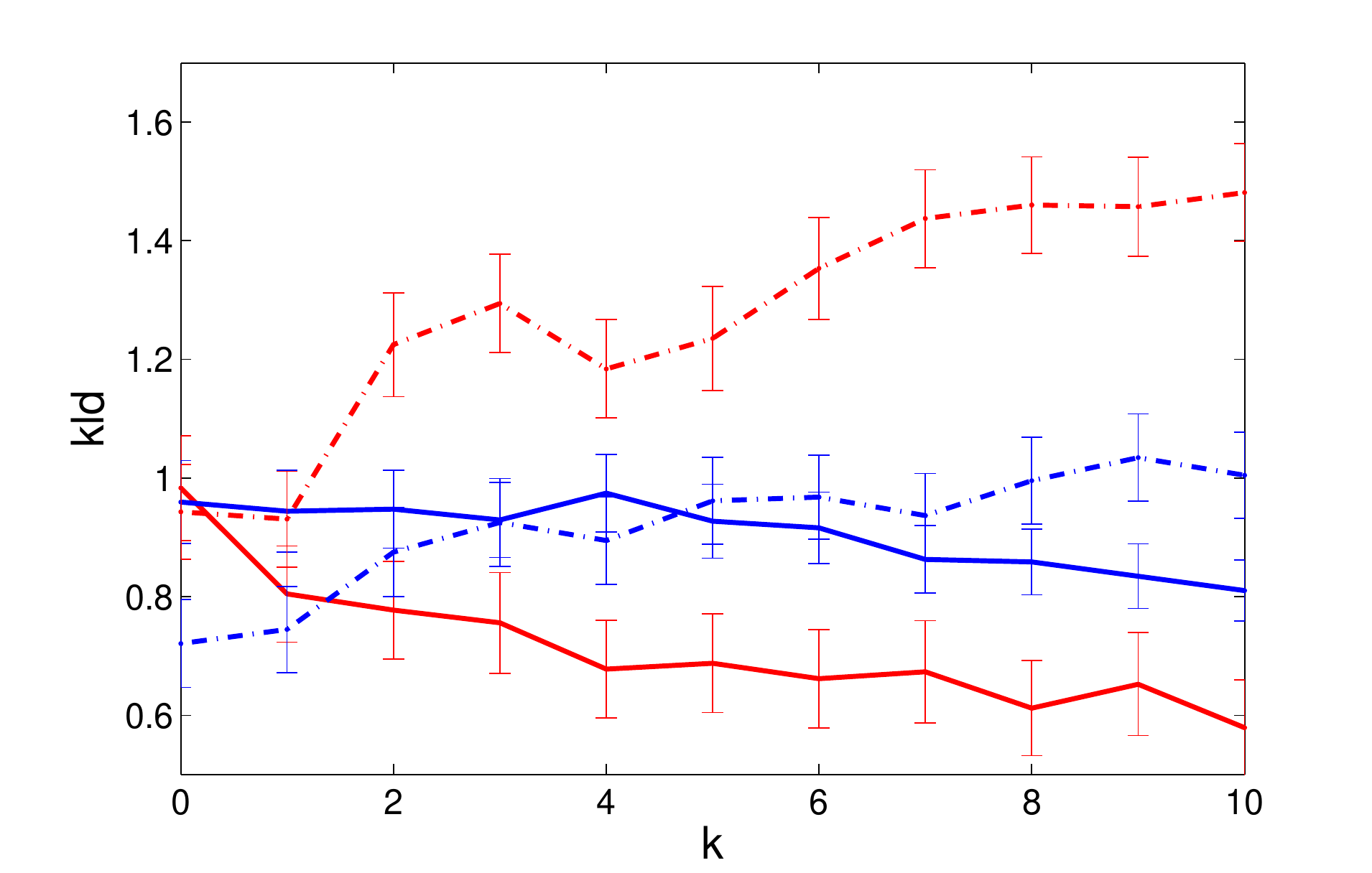} 
\includegraphics[trim = 0mm 1mm 0mm 2mm,clip, width=6cm, height=3.9cm]{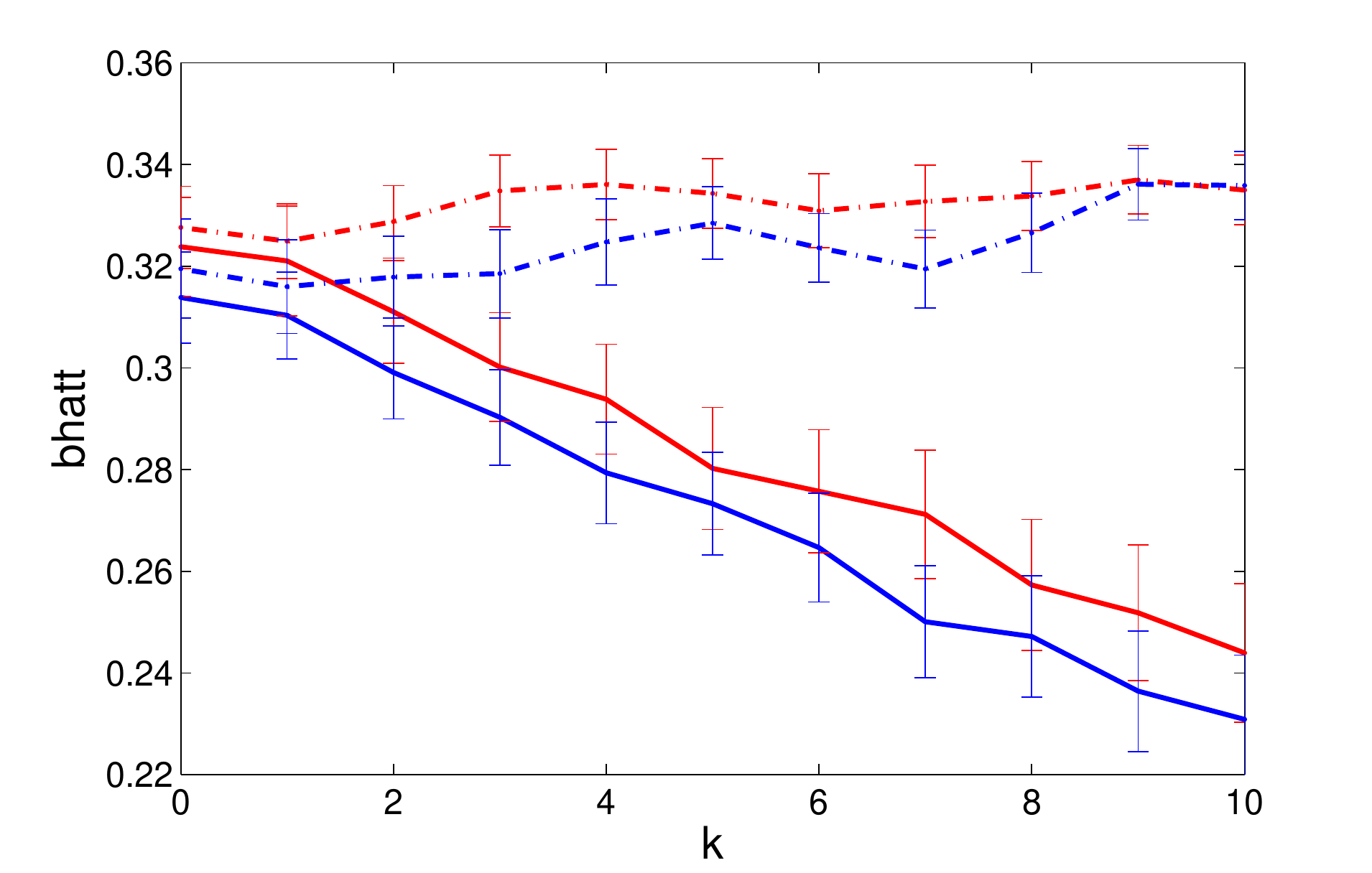} 

\hspace{3cm} (g) \hspace{5cm} (h) \hspace{5cm} (i)

\includegraphics[trim = 0mm 1mm 0mm 2mm,clip, width=6cm, height=3.9cm]{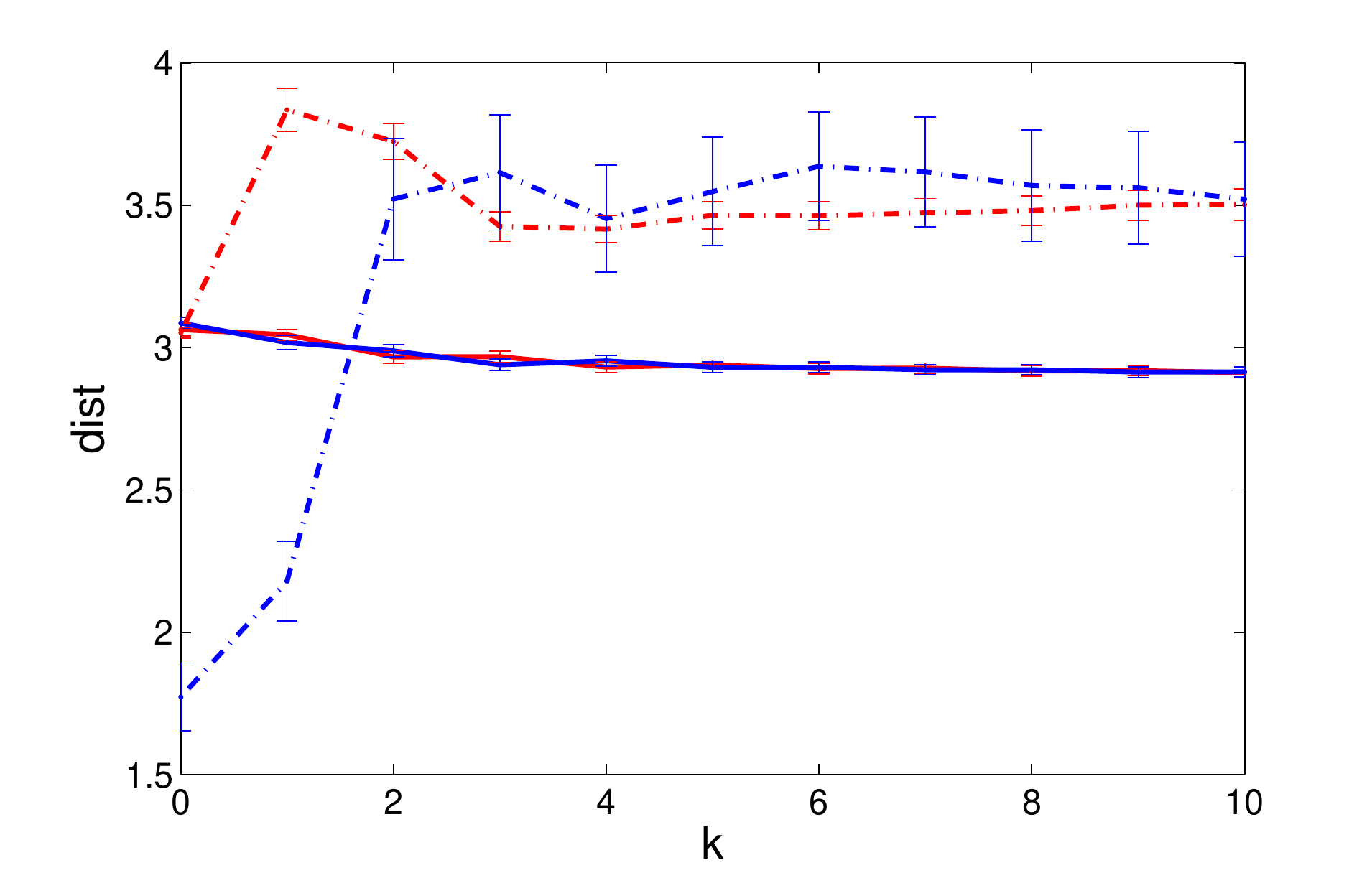} 
\includegraphics[trim = 0mm 1mm 0mm 2mm,clip, width=6cm, height=3.9cm]{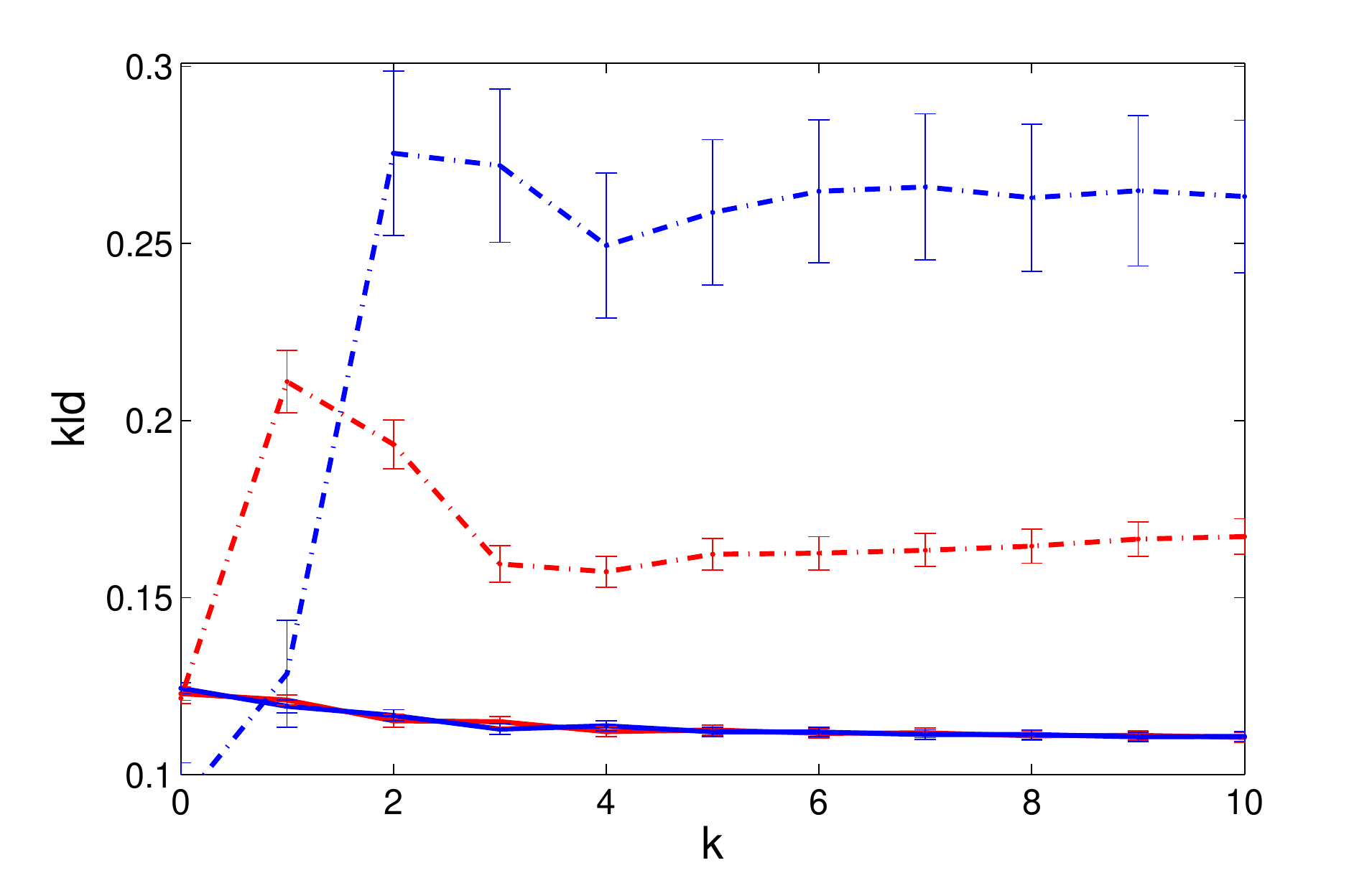} 
\includegraphics[trim = 0mm 1mm 0mm 2mm,clip, width=6cm, height=3.9cm]{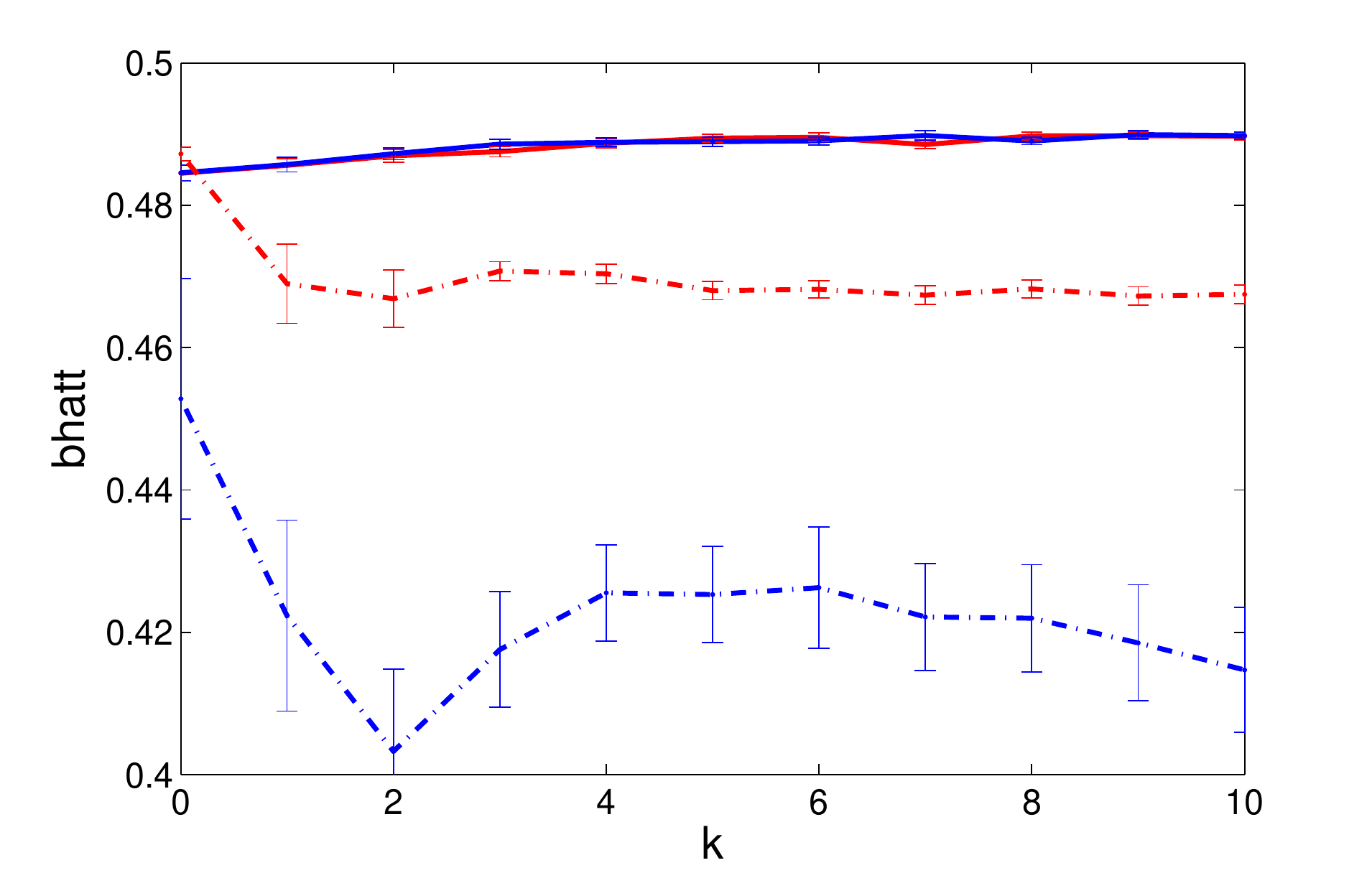} 

\hspace{3cm} j) \hspace{5cm} k) \hspace{5cm} l)

\caption{Landscape measures for the minimal substrates: the test--based problems specified by the objective fitness (\ref{eq:smooth}), (a-c), and  (\ref{eq:crisp}), (d-f). The compositional problem specified by the shared objective fitness function  (\ref{eq:sin}), (g-h), and (\ref{eq:ridge}), (j-l). The  red curves indicate measures over $S_x$, blue curves over $S_y$, solid lines competitive, and dotted lines cooperative interaction, see also the color and line style code in a-c, which applies for all graphs of this figure. }
\label{fig:landmeasures_compo}
\end{figure*}

Fig.  \ref{fig:landmeasures_compo} shows the result for the codynamic landscape measures (\ref{eq:dist}),   (\ref{eq:kld}) and (\ref{eq:bhatt}) for all minimal substrates considered. For the test--based problem specified by the smooth objective fitness function  (\ref{eq:smooth}), refer to Fig.  \ref{fig:landmeasures_compo}a-c and for the piece-wise linear function (\ref{eq:crisp}), see Fig. \ref{fig:landmeasures_compo}d-f. The results of the compositional problem specified by the sinusoid shared objective fitness function  (\ref{eq:sin}) are given in Fig.  \ref{fig:landmeasures_compo}g-i and the landscape measures for the ridge function (\ref{eq:ridge}) are shown in Fig.  \ref{fig:landmeasures_compo}j-l.   As the landscapes measures depend on the outcome of coevolutionary runs, the averages for 100 runs and the $95\%$ confidence intervals are given for coevolutionary generations $k=0$ to $k=10$. Again the red curves indicate the results for the subjective landscape $f_{sub}(x,k)$ of population $P_1$, while the blue curves are for $f_{sub}(y,k)$ of population $P_2$.
A first interesting feature of the two test--based problems (see Fig.  \ref{fig:landmeasures_compo}a-f) is that the measures for the landscapes of cooperating populations are almost indistinguishable. This indicates that the cooperation leads to  landscapes that become very similar. This similarity, however, is only between the two subjective landscapes, but not between subjective and objective landscape. Here, the distance for cooperative interaction is very often larger than for competitive interaction. This is particularly visible for the similarity measure Euclidean distance,  see Fig.  \ref{fig:landmeasures_compo}a,d. The Euclidean distance being larger for cooperative interaction than for competitive interaction  becomes plausible considering the dynamics of the subjective landscapes for the problem (\ref{eq:smooth}). As can be seen in Fig. \ref{fig:landcompcoop}b, the subjective landscapes contract around the solution peak. This contraction is much stronger for cooperative than for competitive interaction. The contraction, on the other hand, also implies a strong deviation from the objective landscape, which in turn means a stronger differences between objective and subjective landscape. This effect is clearly visible in Fig. \ref{fig:landmeasures_compo}a. For the two compositional problems (see Fig.  \ref{fig:landmeasures_compo}g-l), the closeness of the measures for cooperative interaction is also observable, albeit the similarity between the codynamic landscapes is not as strong as for the test--based cases.  Another general features is that for the two compositional problems, the confidence intervals of the measures are generally much larger than for the test--based problems, which implies that the subjective landscapes have a larger variety. Particularly, for the problem   modeled by the ridge function (\ref{eq:ridge}) (Fig.  \ref{fig:landmeasures_compo}j-l) we can see that the landscape over $S_x$ varies much weaker  than the codynamic cooperating landscape over     $S_y$. Two further observations are that the measures do stop to change with run--time after a certain number of coevolutionary generations (indicating that a kind of steady state has been reached), and that the statistical similarity measures (kld and bhatt) largely reflect the geometric measure Euclidean distance (and hence might be usable as an substitute if the geometric measure cannot be calculated), but may also  add further clues for discriminating the codynamics between objective and subjective fitness.

\section{Conclusions}
In this paper an approach has been presented for analyzing coevolution using the theoretical framework of fitness landscapes. It has been shown that the approach can be applied for test--based as well as compositional problems. For these two classes of coevolutionary problems, simple and abstract models, called minimal substrates, were studied for both cooperative as well as competitive interaction.  An important design question in coevolution is whether and under what circumstances subjective fitness implies objective fitness. The dynamic fitness landscape approach aims specifically at addressing this question.  Therefore,  objective and subjective landscapes for the minimal substrates were defined and analyzed. The results have shown that between these landscapes there emerges codynamics where the evolutionary development of one population has effect upon the other population and therefore deforms its subjective fitness landscape. As this process works in both ways the landscapes are coupled and codynamic. We further defined three different landscape measures  that are designed to account for differences between objective and subjective fitness. The numerical results suggest that  these similarity measures  are suitable for quantifying and discriminating the codynamics between objective and subjective fitness.

The results have also shown that the coevolutionary process in one population generates the landscape of the other population and vice versa. In this sense the coevolutionary process creates the landscape in which the process of optimization takes place. As a consequence  a strict separation between problem and problem solving algorithm ceases to exist. The fitness landscape of the problem (objective fitness landscape) still sets the background and framework for (co--)evolutionary dynamics, but how the coevolutionary algorithm perceives the problem is governed by the subjective landscape and how strongly the latter deviates codynamically from the former. Therefore, an important question in analyzing coevolutionary algorithms is what features in the codynamic fitness landscapes the algorithm produces. 
The paper has shown how codynamic landscape measures can be helpful for addressing this question. 
Another value of landscape analysis is that it permits  posing questions of how the properties of the landscape reflect, explain and allow predicting expectable behavior (and possibly performance)  of evolutionary search algorithms. More specifically,  the topology of the landscape can be seen as a predictor of algorithmic behavior. It may be reasonable to assume that these relationships also extend to coevolution.  Thus, for a landscape analysis capable of assessing coevolutionary performance, the similarity measure introduced in this paper should be amended with  topological landscape measures, see e.g.~\cite{mal13} for a recent review. In this context, it might be tempting to assume that the expectable algorithmic performance is best if the average subjective landscape reflects maximally the objective landscape. If this is indeed the case could be studied experimentally using the proposed minimal substrates and recorded performance data, for instance average objective fitness, average subjective fitness, and how both quantities correlate~\cite{dejong07}. These studies could also go along with considering the influence of coevolutionary design 
 parameters, e.g. population size, sample size, mutation strength and rate, and so on.

\end{document}